%% file: main.tex
\documentclass[11pt,a4paper]{article}
\usepackage[hyperref]{acl2019}
\usepackage{times}
\usepackage{latexsym}
\usepackage{booktabs}
\usepackage{hyperref}
\usepackage{url}
\usepackage{xcolor}
\usepackage{todonotes}
\usepackage{wrapfig}
\usepackage{anyfontsize}

\aclfinalcopy 

\title{Can You Tell Me How to Get Past Sesame Street? \\ Sentence-Level Pretraining Beyond Language Modeling}

\author{
Alex Wang,\thanks{\hspace{0.5em}This paper supercedes ``Looking for ELMo's Friends: Sentence-Level Pretraining Beyond Language Modeling'', an earlier version of this work by the same authors. Correspondence to: \texttt{alexwang@nyu.edu}}$~\,^1$ Jan Hula,$^2$ Patrick Xia,$^2$ Raghavendra Pappagari,$^2$ \\
\textbf{R. Thomas McCoy,$^2$ Roma Patel,$^3$ Najoung Kim,$^2$ Ian Tenney,$^4$ Yinghui Huang,$^6$}\\
\textbf{Katherin Yu,$^5$ Shuning Jin,$^7$ Berlin Chen,$^8$ Benjamin Van Durme,$^2$ Edouard Grave,$^5$} \\
\textbf{Ellie Pavlick,$^{3,4}$ and Samuel R. Bowman$^1$}\\
$^1$New York University, $^2$Johns Hopkins University, $^3$Brown University, $^4$Google AI Language,\\$^5$Facebook,
$^6$IBM, $^7$University of Minnesota Duluth, $^8$Swarthmore College
}

\date{}

\begin{document}
\maketitle
\begin{abstract}
Natural language understanding has recently seen a surge of progress with the use of sentence encoders like ELMo \citep{N18-1202} and BERT \citep{devlin2018bert} which are pretrained on variants of language modeling. 
We conduct the first large-scale systematic study of candidate pretraining tasks, comparing 19 different tasks both as alternatives and complements to language modeling.
Our primary results support the use language modeling, especially when combined with pretraining on additional labeled-data tasks.
However, our results are mixed across pretraining tasks and show some concerning trends: In ELMo's pretrain-then-freeze paradigm, random baselines are worryingly strong and results vary strikingly across target tasks. 
In addition, fine-tuning BERT on an intermediate task often negatively impacts downstream transfer. 
In a more positive trend, we see modest gains from multitask training, suggesting the development of more sophisticated multitask and transfer learning techniques as an avenue for further research.
\end{abstract}

\begin{figure}[ht]
\centering
\includegraphics[width=\columnwidth]{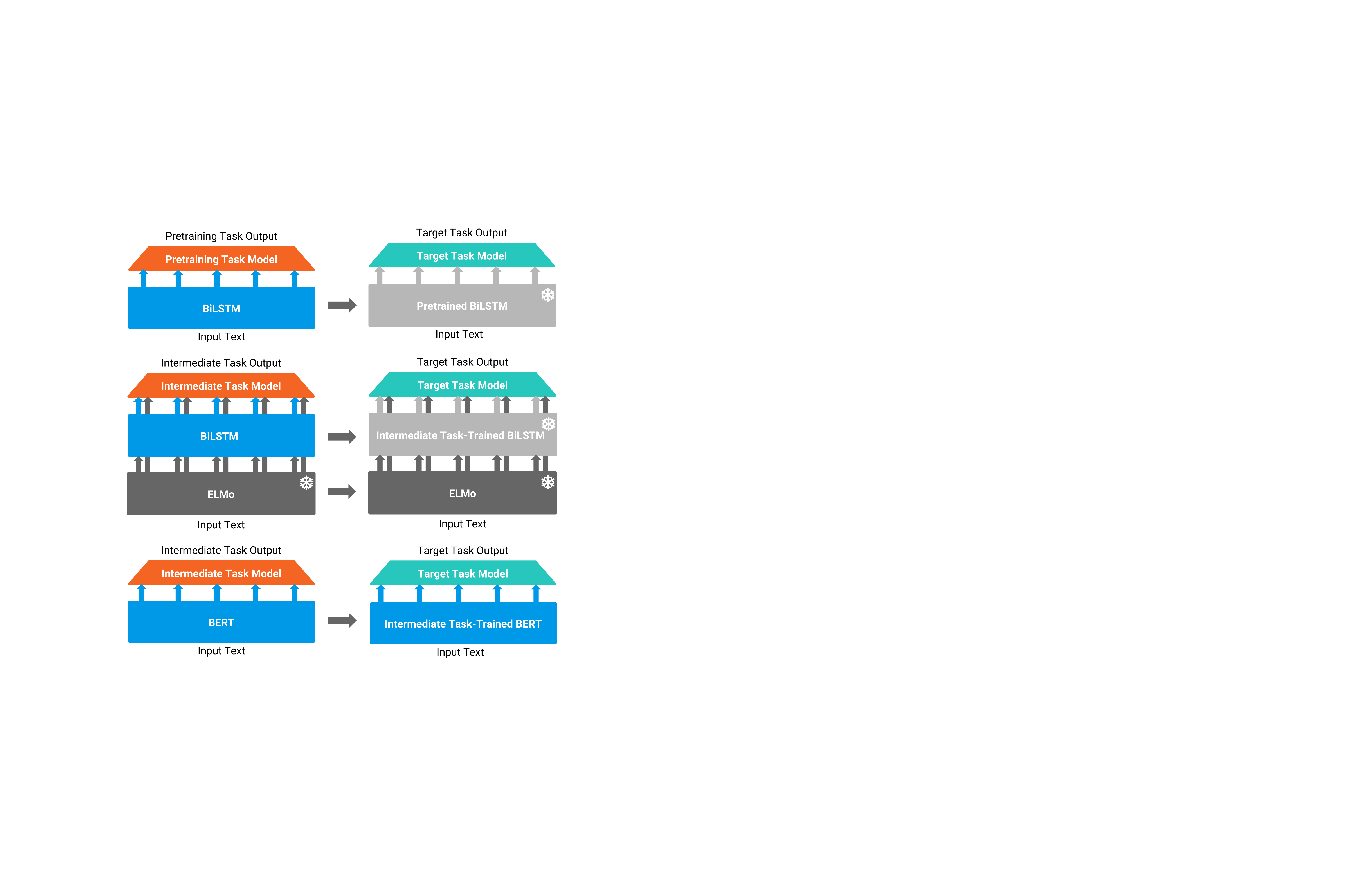}
\caption{\label{fig:schematic} Learning settings that we consider. Model components with frozen parameters are shown in gray and decorated with snowflakes.
\textbf{Top} (pretraining): We pretrain a BiLSTM on a task (left), and learn a target task model on top of the representations it produces (right).
\textbf{Middle} (intermediate ELMo training): We train a BiLSTM on top of ELMo for an intermediate task (left). We then train a target task model on top of the intermediate task BiLSTM and ELMo (right).
\textbf{Bottom} (intermediate BERT training): We fine-tune BERT on an intermediate task (left), and then fine-tune the resulting model again on a target task (right).}
\end{figure}

\section{Introduction}\label{sec:intro}

State-of-the-art models in natural language processing (NLP) often incorporate encoder functions which generate a sequence of vectors intended to represent the in-context meaning of each word in an input text. 
These encoders have typically been trained directly on the target task at hand,
which can be effective for data-rich tasks and yields human performance on some narrowly-defined benchmarks \citep{rajpurkar2018know,achieving-human-parity-on-automatic-chinese-to-english-news-translation}, but is tenable only for the few tasks with millions of training data examples. This limitation has prompted interest in \textit{pretraining} for these encoders: The encoders are first trained on outside data, and then plugged into a target task model.

\citet{P18-1031}, \citet{N18-1202}, \citet{radford2018improving}, and \citet{devlin2018bert} establish that encoders pretrained on variants of the \textit{language modeling} task can be reused to yield strong performance on downstream NLP tasks.
Subsequent work has homed in on language modeling (LM) pretraining, finding that such models can be productively fine-tuned on \textit{intermediate} tasks
like natural language inference 
before transferring to downstream tasks \citep{phang2018sentence}.
However, we identify two open questions: 
(1) How effective are tasks beyond language modeling in training reusable sentence encoders 
(2) Given the recent successes of LMs with intermediate-task training, which tasks can be effectively combined with language modeling and each other.

The main contribution of this paper is a large-scale systematic study of these two questions.
For the first question, we train reusable sentence encoders on 19 different pretraining tasks and task combinations and several simple baselines, using a standardized model architecture and procedure for pretraining.
For the second question, we conduct additional pretraining on ELMo \citep{peters2018dissecting} and BERT \citep{devlin2018bert} with 17 different intermediate tasks and task combinations.
We evaluate each of these encoders on the nine target language-understanding tasks in the GLUE benchmark \citep{wang2018glue}, yielding a total of 53 sentence encoders and 477 total trained models. 
We measure correlation in performance across target tasks and plot learning curves to show the effect of data volume on both pretraining and target task training. 

We find that language modeling is the most effective pretraining task that we study. 
Multitask pretraining or intermediate task training offers modest further gains.
However, we see several worrying trends:
\begin{itemize}
\item The margins between substantially different pretraining tasks can be extremely small in this transfer learning regimen and many pretraining tasks struggle to outperform trivial baselines.
\item Many of the tasks used for intermediate task training \textit{adversely} impact the transfer ability of LM pretraining.
\item Different target tasks differ dramatically in what kinds of pretraining they benefit most from, but na\"ive multitask pretraining seems ineffective at combining the strengths of disparate pretraining tasks.
\end{itemize}
These observations suggest that 
while scaling up LM pretraining \citep[as in][]{radford2019language} is likely the most straightforward path to further gains, 
our current methods for multitask and transfer learning may be substantially limiting our results.

\section{Related Work}\label{sec:related}

Work on reusable sentence encoders can be traced back at least as far as the multitask model of \citet{collobert2011natural}.  
Several works focused on learning reusable \textit{sentence-to-vector} encodings, where the pretrained encoder produces a fixed-size representation for each input sentence \citep{dai2015semi,kiros2015skip,hill2016learning,DBLP:conf/emnlp/ConneauKSBB17}.
More recent reusable sentence encoders such as CoVe \citep{mccann2017learned} and GPT \citep{radford2018improving} instead represent sentences as \textit{sequences} of vectors. 
These methods work well, but most use distinct pretraining objectives, and none offers a substantial investigation of the choice of objective like we conduct here.

We build on two methods for pretraining sentence encoders on language modeling: ELMo and BERT. 
ELMo consists of a forward and backward LSTM \citep{hochreiter1997long}, the hidden states of which are used to produce a contextual vector representation for each token in the inputted sequence. 
ELMo is adapted to target tasks by freezing the model weights and only learning a set of task-specific scalar weights that are used to compute a linear combination of the LSTM layers.
BERT consists of a pretrained Transformer \citep{vaswani2017attention}, 
and is adapted to downstream tasks by fine-tuning the entire model. Follow-up work has explored parameter-efficient fine-tuning \citep{stickl2019bert,houlsby2019parameter} and better target task adaptation via multitask fine-tuning \citep{phang2018sentence,liu2019multitask}, but work in this area is nascent.

The successes of sentence encoder pretraining have sparked a line of work analyzing these models  \citep[][i.a.]{zhang2018lessons,peters2018dissecting,tenney2018what,peters2019tune,tenney2019bert,liu2019multitask}. 
Our work also attempts to better understand what is learned by pretrained encoders, but we study this question entirely through the lens of pretraining and fine-tuning tasks, 
rather than architectures or specific linguistic capabilities.
Some of our experiments resemble those of \citet{yogatama2019learning}, who also empirically investigate transfer performance with limited amounts of data and find similar evidence of catastrophic forgetting.

Multitask representation learning in NLP is well studied, and again can be traced back at least as far as \citet{collobert2011natural}. \citet{luong2015multi} show promising results combining translation and parsing; \citet{subramanian2018large} benefit from multitask learning in sentence-to-vector encoding; and \citet{bingel-sogaard:2017:EACLshort} and \citet{changpinyo-hu-sha:2018:C18-1} offer studies of when multitask learning is helpful for lower-level NLP tasks.

\section{Transfer Paradigms}\label{sec:transfer}

We consider two recent paradigms for transfer learning: pretraining and intermediate training. See Figure~\ref{fig:schematic} for a graphical depiction.

\paragraph{Pretraining}
Our first set of experiments is designed to systematically investigate the effectiveness of a broad range of tasks in pretraining sentence encoders.
For each task, we first train a randomly initialized model to convergence on that \textit{pretraining task}, and then train a model for a \textit{target task} on top of the trained encoder.
For these experiments, we largely follow the procedure and architecture used by ELMo rather than BERT, but we expect similar trends with BERT-style models.

\paragraph{Intermediate Training}

Given the robust success of LM pretraining, we explore methods of further improving on such sentence encoders.
In particular, we take inspiration from \citet{phang2018sentence}, who show gains in first fine-tuning BERT on an \textit{intermediate task}, and then fine-tuning again on a target task.
Our second set of experiments investigates which tasks can be used for intermediate training to augment LM pretraining. 
We design experiments using both pretrained ELMo and BERT as the base encoder. 
When using ELMo, we follow standard procedure and train a task-specific LSTM and output component (e.g. MLP for classification, decoder for sequence generation, etc.) on top of the representations produced by ELMo.
During this stage, the pretrained ELMo weights are frozen except for a set of layer mixing weights.
When using BERT, we follow standard procedure and train a small task-specific output component using the \texttt{[CLS]} output vector while also fine-tuning the weights of the full BERT model.

\paragraph{Target Task Evaluation}

For our pretraining and intermediate ELMo experiments, to evaluate on a target task, we train a target task model on top of the representations produced by the encoder, which is again frozen throughout target task training except for a set of target-task-specific layer mixing weights.
For our intermediate BERT experiments, we follow the same procedure as in intermediate training:
We train a target-task model using the \texttt{[CLS]} representation and fine-tune the encoder throughout target task training.

We use the nine target tasks in GLUE \citep{wang2018glue} to evaluate each of the encoders we train.
GLUE is an open-ended shared task competition and evaluation toolkit for reusable sentence encoders, built around a set of nine sentence and sentence pairs tasks spanning a range of dataset sizes, paired with private test data and an online leaderboard.
We evaluate each model on each of the nine tasks, and report the resulting scores and the GLUE score, a macro-average over tasks.

\section{Tasks}\label{sec:tasks}

\begin{table}[t]
\centering \fontsize{8.4}{10.1}\selectfont \setlength{\tabcolsep}{0.5em}
\input{tables/tasks.tex}
\caption{Tasks used for pretraining and intermediate training of sentence encoders. We also use the GLUE tasks as target tasks to evaluate the encoders. For the language modeling (LM) tasks, we report the number of sentences in the corpora.}
\label{tab:tasks}
\end{table}

Our experiments compare encoders pretrained or fine-tuned on a large number of tasks and task combinations, where a \textit{task} is a dataset--objective function pair. 
We select these tasks either to serve as baselines or because they have shown promise in prior work, especially in sentence-to-vector encoding. 
See Appendix \ref{ax:per-task} for details and tasks we experimented with but which did not show strong enough performance to warrant a full evaluation.

\paragraph{Random Encoder} 

A number of recent works have noted that randomly initialized, untrained LSTMs can obtain surprisingly strong downstream task performance \cite{zhang2018lessons,wieting2019training,tenney2018what}.
Accordingly, our pretraining and intermediate ELMo experiments include a baseline of a randomly initialized BiLSTM with no further training.
This baseline is especially strong because our ELMo-style models use a skip connection from the input of the encoder to the output, allowing the task-specific component to see the input representations, yielding a model similar to \citet{iyyer2015deep}.

\paragraph{GLUE Tasks}

We use the nine tasks included with GLUE as pretraining and intermediate tasks:
acceptability classification with \textbf{CoLA} \citep{warstadt2018neural}; binary sentiment classification with \textbf{SST} \citep{socher2013recursive}; semantic similarity with the MSR Paraphrase Corpus \citep[\textbf{MRPC};][]{dolan2005automatically}, Quora Question Pairs\footnote{ \href{https://data.quora.com/First-Quora-Dataset-Release-Question-Pairs}{\texttt{data.quora.com/\allowbreak First-\allowbreak Quora-\allowbreak Dataset-\allowbreak Release-Question-Pairs}}} (\textbf{QQP}), and STS-Benchmark \citep[\textbf{STS};][]{cer2017semeval}; and textual entailment with the Multi-Genre NLI Corpus \citep[\textbf{MNLI}][]{DBLP:journals/corr/WilliamsNB17}, RTE 1, 2, 3, and 5 \citep[\textbf{RTE};][et seq.]{dagan2006pascal}, and data from SQuAD \citep[\textbf{QNLI};\footnote{QNLI has been re-released with updated splits since the original release. We use the original splits.}][]{rajpurkar2016squad} and the Winograd Schema Challenge \citep[\textbf{WNLI};][]{levesque2011winograd} recast as entailment in the style of \citet{white2017inference}. 
MNLI and QQP have previously been shown to be effective for pretraining in other settings \citep{DBLP:conf/emnlp/ConneauKSBB17,subramanian2018large,phang2018sentence}.
Other tasks are included to represent a broad sample of labeling schemes commonly used in NLP.

\paragraph{Outside Tasks}
 
We train language models on two datasets: WikiText-103 \citep[\textbf{WT};][]{DBLP:journals/corr/MerityXBS16} and Billion Word Language Model Benchmark \citep[\textbf{BWB};][]{chelba2013one}.
Because representations from ELMo and BERT capture left and right context, they cannot be used in conjunction with unidirectional language modeling, so we exclude this task from intermediate training experiments.
We train machine translation (MT) models on \textbf{WMT14 English-German} \citep{bojar2014findings} and \textbf{WMT17 English-Russian} \citep{W17-4700}. 
We train \textbf{SkipThought}-style sequence-to-sequence (seq2seq) models to read a sentence from WT and predict the following sentence
\citep{kiros2015skip,W17-2625}.
We train \textbf{DisSent} models to read two clauses from WT that are connected by a discourse marker such as \textit{and}, \textit{but}, or \textit{so} and predict the the discourse marker \citep{jernite2017discourse,nie2017dissent}.
Finally, we train seq2seq models to predict the response to a given comment from \textbf{Reddit}, using a previously existing dataset obtained by a third party (available on pushshift.io), comprised of 18M comment--response pairs from 2008-2011. This dataset was used by \citet{yang2018learning} to train sentence encoders.

\paragraph{Multitask Learning}

We consider three sets of these tasks for multitask pretraining and intermediate training: all GLUE tasks, all non-GLUE (outside) tasks, and all tasks.

\section{Models and Experimental Details}\label{sec:models}

We implement our models using the \texttt{jiant} toolkit,\footnote{\url{https://github.com/nyu-mll/jiant/tree/bert-friends-exps}} 
which is in turn built on AllenNLP \citep{Gardner2017AllenNLP} and on a public PyTorch implementation of BERT.\footnote{\url{https://github.com/huggingface/pytorch-pretrained-BERT}}
Appendix~\ref{ax:per-task} presents additional details.

\paragraph{Encoder Architecture}

For both the pretraining and intermediate ELMo experiments, we process words using a pretrained character-level convolutional neural network (CNN) from ELMo. 
We use this pretrained word encoder for pretraining experiments to avoid potentially difficult issues with unknown word handling in transfer learning.

For the pretraining experiments, these input representations are fed to a two-layer 1024D bidirectional LSTM from which we take the sequence of hidden states from the top layer as the contextual representation.
A task-specific model sees both the top-layer hidden states of this model and, through a skip connection, the input token representations.
For the intermediate ELMo experiments, we compute contextual representations using the entire pretrained ELMo model, which are passed to a similar LSTM that is then trained on the intermediate task. 
We also include a skip connection from the ELMo representations to the task specific model.
Our experiments with BERT use the \textsc{base} case-sensitive version of the model.

\paragraph{Task-Specific Components}

\begin{table*}[t]
\centering \fontsize{8.4}{10.1}\selectfont \setlength{\tabcolsep}{0.5em}
\input{tables/pretraining.tex}
\caption{Results for \textbf{pretraining} experiments on development sets except where noted. 
\textbf{Bold} denotes best result overall.
\underline{Underlining} denotes an average score surpassing the \textit{Random} baseline.
See Section \ref{sec:results} for discussion of WNLI results (*).
}
\label{tab:glueresults}
\end{table*}

We design task-specific components to be as close to standard models for each task as possible. 
Though different components may have varying parameter counts, architectures, etc., we believe that results between tasks are still comparable and informative.

For BERT experiments we use the standard preprocessing and pass the representation of the special \texttt{[CLS]} representation to a logistic regression classifier. For seq2seq tasks (MT, SkipThought, pushshift.io Reddit dataset) we replace the classifier with a single-layer LSTM word-level decoder and initialize the hidden state with the \texttt{[CLS]} representation.


For ELMo-style models, we use several model types:
\begin{itemize}
\item \textbf{Single-sentence classification tasks}: We train a linear projection over the output states of the encoder, max-pool those projected states, and feed the result to an MLP. 

\item \textbf{Sentence-pair tasks}: We perform the same steps on both sentences and use the heuristic feature vector $[h_1; h_2; h_1 \cdot h_2; h_1 - h_2]$ in the MLP, following \citet{mou-EtAl:2016:P16-2}. 
When training target-task models on QQP, STS, MNLI, and QNLI, 
we use a cross-sentence attention mechanism similar to BiDAF \citep{seo2016bidirectional}.
We do not use this mechanism in other cases as early results indicated it hurt transfer performance. 

\item \textbf{Seq2seq tasks} (MT, SkipThought, pushshift.io Reddit dataset): We use a single-layer LSTM decoder where the hidden state is initialized with the pooled input representation.

\item \textbf{Language modeling}: We follow ELMo by concatenating forward and backward models and learning layer mixing weights.
\end{itemize}

To use GLUE tasks for pretraining or intermediate training in a way that is more comparable to outside tasks,
after pretraining we discard the learned GLUE classifier, and initialize a new classifier from scratch for target-task training.

\begin{table*}[h!]
\centering \fontsize{8.4}{10.1}\selectfont \setlength{\tabcolsep}{0.5em}
\input{tables/intermediate-by-model.tex}
\caption{Results for \textbf{intermediate training} experiments on development sets except where noted. $^E$ and $^B$ respectively denote ELMo and BERT experiments.
\textbf{Bold} denotes best scores by section.
\underline{Underlining} denotes average scores better than the single-task baseline.
See Section \ref{sec:results} for discussion of WNLI results (*).
BERT Base numbers are from \citet{devlin2018bert}.
}
\label{tab:intermediate}
\end{table*}

\paragraph{Training and Optimization} 

For BERT experiments, we train our models with the same optimizer and learning rate schedule as the original work. For all other models, we train our models with AMSGrad \citep[][]{j.2018on}. 
We do early stopping using development set performance of the task we are training on. 
Typical experiments (pretraining or intermediate training of an encoder and training nine associated target-task models) take 1--5 days to complete on an NVIDIA P100 GPU. 

When training on multiple tasks, we randomly sample a task with probability proportional to its training data size raised to the power of 0.75. This sampling rate is meant to balance the risks of overfitting small-data tasks and underfitting large ones, and performed best in early experiments. More extensive experiments with methods like this are shown in Appendix~\ref{ax:mtl}. We perform early stopping based on an average of the tasks' validation metrics.

\paragraph{Hyperparameters} \label{sec:hyper} Appendix~\ref{ax:hyper} lists the hyperparameter values used. As our experiments require more than 150 GPU-days on NVIDIA P100 GPUs to run---not counting debugging or learning curves---we do not have the resources for extensive tuning. 
Instead, we fix most hyperparameters to commonly used values.
The lack of tuning limits our ability to diagnose the causes of poor performance when it occurs, and we invite readers to further refine our models using the public code.

\begin{figure*}[t]
\centering
\includegraphics[width=0.95\linewidth]{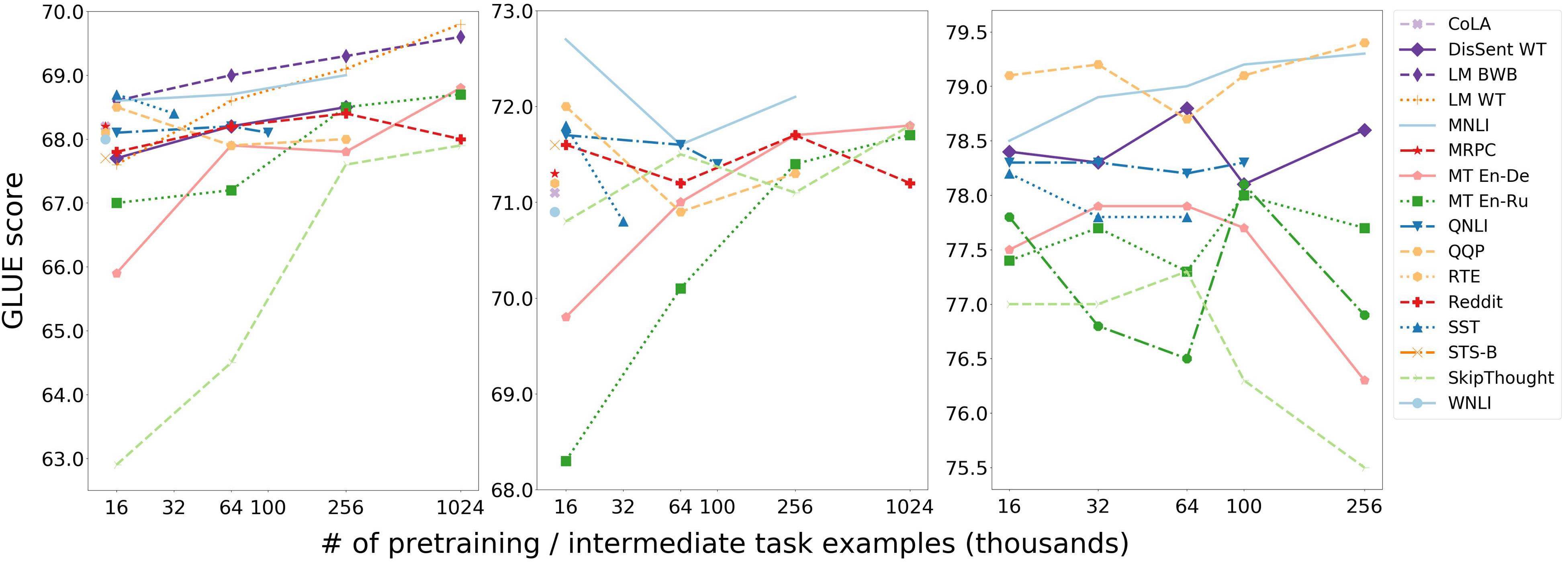}
\caption{Learning curves (log scale) showing overall GLUE scores for encoders pretrained to convergence with varying amounts of data, shown for pretraining (left) and intermediate ELMo (center) and BERT (right) training.
\label{fig:lc}}
\end{figure*}

\section{Results}\label{sec:results}

Tables~\ref{tab:glueresults} and \ref{tab:intermediate} respectively show results for our pretraining and intermediate training experiments.
The \textit{Single-Task} baselines train and evaluate a model on only the corresponding GLUE task.
To comply with GLUE's limits on test set access, we only evaluate the top few pretrained encoders. 
For roughly comparable results in prior work, see \citet{wang2018glue} or \url{www.gluebenchmark.com}; we omit them here in the interest of space. 
As of writing, the best test result using a comparable frozen pretrained encoder is 70.0 from \citet{wang2018glue} for a model similar to our GLUE$^E$, and the best overall published result is 85.2 from \citet{liu2019multitask} using a model similar to our GLUE$^B$ (below), but substantially larger.

While it is not feasible to run each setting multiple times, we estimate the variance of the GLUE score by re-running three experiments five times each with different random seeds. 
We observe $\sigma=0.4$ for the random encoder with no pretraining, $\sigma=0.2$ for ELMo with intermediate MNLI training, and $\sigma=0.5$ for BERT without intermediate training.
This variation is substantial but many of our results surpass a standard deviation of our baselines.

The WNLI dataset is both difficult and  adversarial: The same hypotheses can be paired with different premises and opposite labels in the train and development sets, so models that overfit the train set (which happens quickly on the tiny training set) often show development set performance below chance, making early stopping and model selection difficult.
Few of our models reached even the \textit{most frequent class} performance (56.3), and when evaluating models that do worse than this, we replace their predictions with the most frequent label to simulate the performance achieved by not modeling the task at all.

\subsection{Pretraining}

From Table~\ref{tab:glueresults}, among target tasks, we find the grammar-related CoLA task benefits dramatically from LM pretraining: 
The results achieved with LM pretraining are significantly better than the results achieved without.
In contrast, the meaning-oriented 
STS sees good results with several kinds of pretraining, but does not benefit substantially from LM pretraining.

Among pretraining tasks, language modeling performs best, followed by MNLI. The remaining pretraining tasks yield performance near that of the random baseline.
Even our single-task baseline gets less than a one point gain over this simple baseline.
The multitask models are tied or outperformed by models trained on one of their constituent tasks, suggesting that our approach to multitask learning does not reliably produce models that productively combine the knowledge taught by each task. 
However, of the two models that perform best on the development data, the multitask model generalizes better than the single-task model on test data for tasks like STS and MNLI where the test set contains out-of-domain data.

\paragraph{Intermediate Task Training}

Looking to Table \ref{tab:intermediate},
using ELMo uniformly improves over training the encoder from scratch.
The ELMo-augmented random baseline is strong, lagging behind the single-task baseline by less than a point.
Most intermediate tasks beat the random baseline, but several fail to significantly outperform the single-task baseline.
MNLI and English--German translation perform best with ELMo, with SkipThought and DisSent also beating the single-task baseline. 
Intermediate multitask training on all the non-GLUE tasks produces our best-performing ELMo model.

Using BERT consistently outperforms ELMo and pretraining from scratch.
We find that intermediate training on each of MNLI, QQP, and STS leads to improvements over 
no intermediate training,
while intermediate training on the other tasks harms transfer performance. 
The improvements gained via STS, a small-data task, versus the negative impact of fairly large-data tasks (e.g. QNLI), suggests that the benefit of intermediate training is not solely due to additional training, but that the signal provided by the intermediate task complements the original language modeling objective. 
Intermediate training on generation tasks such as MT and SkipThought significantly impairs BERT's transfer ability. 
We speculate that this degradation may be due to catastrophic forgetting in fine-tuning for a task substantially different from the tasks BERT was originally trained on.
This phenomenon might be mitigated in our ELMo models via the frozen encoder and skip connection.
On the test set, we lag slightly behind the BERT base results from \citet{devlin2018bert}, likely due in part to our limited hyperparameter tuning.

\begin{table}[t]
\centering \fontsize{8.0}{10}\selectfont \setlength{\tabcolsep}{0.5em}
\input{tables/correlations-pretrain-subset.tex}
\caption{\label{tab:corr}Pearson correlations between performances on a subset of all pairs of target tasks, measured over all runs reported in Table \ref{tab:glueresults}. The \textit{Avg} column shows the correlation between performance on a target task and the overall GLUE score.
For QQP and STS, the correlations are computed respectively using F1 and Pearson correlation.
Negative correlations are \underline{underlined}.}
\end{table}

\section{Analysis and Discussion}\label{sec:analysis}

\paragraph{Target Task Correlations}

Table~\ref{tab:corr} presents an alternative view of the results of the pretraining experiment (Table~\ref{tab:glueresults}): 
The table shows correlations between pairs of target tasks over the space of pretrained encoders. The correlations reflect the degree to which the performance on one target task with some encoder predicts performance on another target task with the same encoder.
See Appendix \ref{ax:correlations} for the full table and similar tables for intermediate ELMo and BERT experiments.

Many correlations are low, suggesting that different tasks benefit from different forms of pretraining to a substantial degree, and bolstering the observation that no single pretraining task yields good performance on all target tasks. For reasons noted earlier, the models that tended to perform best overall also tended to overfit the WNLI training set most, leading to a negative correlation between WNLI and overall GLUE score. 
STS also shows a negative correlation, likely due to the observation that it does not benefit from LM pretraining. 
In contrast, CoLA shows a strong correlation with the overall GLUE scores, but has weak or negative correlations with many tasks: The use of LM pretraining dramatically improves CoLA performance, but most other forms of pretraining have little effect.

\paragraph{Learning Curves}

Figure~\ref{fig:lc} shows performance on the overall GLUE metric for encoders pretrained to convergence on each task with varying amounts of data. 
Looking at pretraining tasks in isolation (left), most tasks improve slightly as the amount of data increases, with the LM and MT tasks showing the most promising combination of slope and maximum performance. 
Combining these tasks with ELMo (center) or BERT (right) yields less interpretable results: the relationship between training data volume and performance becomes weaker, and some of the best results reported in this paper are achieved by models that combine ELMo with \textit{restricted-data versions} of intermediate tasks like MNLI and QQP.
This effect is amplified with BERT, with training data volume having unclear or negative relationships with performance for many tasks.
With large datasets for generation tasks, we see clear evidence of catastrophic forgetting with performance sharply decreasing in amount of training data.

We also measure the performance of target task performance for three fully pretrained encoders under varying amounts of target task data.
We find that all tasks benefit from increasing data quantities, with no obvious diminishing returns, and that most tasks see a consistent improvement in performance with the use of pretraining, regardless of the data volume.
We present these learning curves in Appendix \ref{ax:curves}.

\paragraph{Results on the GLUE Diagnostic Set}
On GLUE's analysis dataset, we find that many of our pretraining tasks help on examples involving lexical-semantic knowledge and logical operations, but less so on examples that highlight world knowledge. See Appendix \ref{ax:diag} for details. 

\section{Conclusions}

We present a systematic comparison of tasks and task combinations for the pretraining and intermediate fine-tuning of sentence-level encoders like those seen in ELMo and BERT. With nearly 60 pretraining tasks and task combinations and nine target tasks, this represents a far more comprehensive study than any seen on this problem to date.

Our primary results are perhaps unsurprising: LM works well as a pretraining task, and no other single task is consistently better. 
Intermediate training of language models can yield modest further gains.
Multitask pretraining can produce results better than any single task can.
Target task performance continues to improve with more LM data, even at large scales, suggesting that further work scaling up LM pretraining is warranted.

We also observe several worrying trends. 
Target tasks differ significantly in the pretraining tasks they benefit from, with correlations between target tasks often low or negative.
Multitask pretraining fails to reliably produce models better than their best individual components.
When trained on intermediate tasks like MT that are highly different than its original training task, BERT shows signs of catastrophic forgetting. 
These trends suggest that improving on LM pretraining with current techniques will be challenging.

While further work on language modeling seems straightforward and worthwhile, we believe that the future of this line of work will require a better understanding of the settings in which target task models can effectively utilize outside knowledge and data, and new methods for pretraining and transfer learning to do so.

\section*{Acknowledgments}
Parts of this work were conducted as part of the Fifth Frederick Jelinek Memorial Summer Workshop (JSALT) at Johns Hopkins University, and benefited from support by the JSALT sponsors and a team-specific donation of computing resources from Google. We gratefully acknowledge the support of NVIDIA Corporation with the donation of a Titan V GPU used at NYU for this research. 
AW is supported by the National Science Foundation Graduate Research Fellowship Program under Grant No. DGE 1342536. PX and BVD were supported by DARPA AIDA. Any opinions, findings, and conclusions or recommendations expressed in this material are those of the authors and do not necessarily reflect the views of the National Science Foundation.

\bibliography{acl2019}
\bibliographystyle{acl_natbib}

\newpage
\appendix

\begin{table}[t]
\centering \fontsize{8.4}{10.1}\selectfont \setlength{\tabcolsep}{0.5em}
\begin{tabular}{lrrrr}
\toprule
 & \bf Small & \bf Medium & \bf Large                         \\
\midrule
Steps btw. validations            & 100               & 100              & 1000                   \\
Attention                         & N                 & N                & Y                     \\
Classifier dropout rate           & 0.4               & 0.2              & 0.2                    \\
Classifier hidden dim.            & 128               & 256              & 512                    \\
Max pool projection dim. & 128               & 256              & 512      \\
\bottomrule
\end{tabular}
\caption{Hyperparameter settings for target-task models and target-task training for ELMo-style models. Small-data tasks are RTE and WNLI; medium-data tasks are CoLA, SST, and MRPC; large-data tasks are STS, QQP, MNLI, and QNLI. STS has a relatively small training set, but consistently patterns with the larger tasks in its behavior. \label{tab:hypertarget}}
\end{table}

\section{Additional Pretraining Task Details}\label{ax:per-task}

\paragraph{DisSent} 

To extract discourse model examples from the WikiText-103 corpus \citep{DBLP:journals/corr/MerityXBS16}, we follow the procedure described in \newcite{nie2017dissent} by extracting clause-pairs that follow specific dependency relationships within the corpus  \citep[see Figure 4 in][]{nie2017dissent}. We use the Stanford Parser \citep{chen2014fast} distributed in Stanford CoreNLP version 3.9.1 to identify the relevant dependency arcs.



\paragraph{Cross-Sentence Attention}

For MNLI, QQP, QNLI, and STS with ELMo-style models, we use an attention mechanism between all pairs of words representations, followed by a $512D\times2$ BiLSTM with max-pooling over time, following the mechanism used in BiDAF \citep{seo2016bidirectional}.

\paragraph{Alternative Tasks} Any large-scale comparison like the one attempted in this paper is inevitably incomplete. Among the thousands of publicly available NLP datasets, we also performed initial trial experiments on several datasets for which we were not able to reach development-set performance above that of the random encoder baseline in the pretraining or as an intermediate task with ELMo. These include image-caption matching with MSCOCO \citep{lin2014microsoft}, following \citet{kiela2017learning};
the small-to-medium-data text-understanding tasks collected in NLI format by \citet{poliak2018towards}; ordinal common sense inference \citep{zhang2017ordinal}; POS tagging on the Penn Treebank \citep{PTB}; supertagging on CCGBank \citep{CCG}; and a variant objective on our Reddit data, inspired by \citet{yang2018learning}, where the model is trained to select which of two candidate replies to a given comment is correct.

\section{Hyperparameters and Optimization Details}\label{ax:hyper} 

See Section~\ref{sec:hyper} for general comments on hyperparameter tuning.

\paragraph{Validation} We evaluate on the validation set for the current training task or tasks every 1,000 steps, except where noted otherwise for small-data target tasks. During multitask learning, we multiply this interval by the number of tasks, evaluating every 9,000 steps during GLUE multitask training, for example. 

\paragraph{Optimizer} For BERT, we use the same optimizer and learning rate schedule as \citet{devlin2018bert}, with an initial learning rate of 1e-5 and training for a maximum of three epochs at each stage (or earlier if we trigger a different early stopping criterion).
For all other experiments, we use AMSGrad \citep{j.2018on}. During pretraining, we use a learning rate of 1e-4 for classification and regression tasks, and 1e-3 for text generation tasks. During target-task training, we use a learning rate of 3e-4 for all tasks.

\paragraph{Learning Rate Decay} We multiply the learning rate by 0.5 whenever validation performance fails to improve for more than 4 validation checks. We stop training if the learning rate falls below 1e-6.

\paragraph{Early Stopping} We maintain a saved checkpoint reflecting the best validation result seen so far. We stop training if we see no improvement after more than 20 validation checks. After training, we use the last saved checkpoint.

\paragraph{Regularization} For BERT models, we follow the original work. For non-BERT models, we apply dropout with a drop rate of 0.2 after the character CNN in pretraining experiments or after ELMo, after each LSTM layer, and after each MLP layer in the task-specific classifier or regressor. For small-data target tasks, we increase MLP dropout to 0.4 during target-task training.

\paragraph{Preprocessing} For BERT, we follow \citet{devlin2018bert} and use the WordPiece \citep{wu2016google} tokenizer. For all other experiments, we use the Moses tokenizer for encoder inputs, and set a maximum sequence length of 40 tokens. There is no input vocabulary, as we use ELMo's character-based input layer. 

For English text generation tasks, we use the Moses tokenizer to tokenize our data, 
but use a word-level output vocabulary of 20,000 types for tasks that require text generation. For translation tasks, we use BPE tokenization with a vocabulary of 20,000 types. For all sequence-to-sequence tasks we train word embeddings on the decoder side. 

\paragraph{Target-Task-Specific Parameters}

For non-BERT models, to ensure that baseline performance for each target task is competitive, we find it necessary to use slightly different models and training regimes for larger and smaller target tasks. We used partially-heuristic tuning to separate GLUE tasks into \mbox{big-,} medium- and small-data groups, giving each group its own heuristically chosen task-specific model specifications. Exact values are shown in Table~\ref{tab:hypertarget}.

\paragraph{Sequence-to-Sequence Models} We found bilinear attention to be helpful for the SkipThought and Reddit pretraining tasks but not for machine translation, and report results for these configurations. For ELMo-style models, we use the max-pooled output of the encoder to initialize the hidden state of the decoder, and the size of this hidden state is equal to the size of the output of our shared encoder. For BERT, we use the representation corresponding to the \texttt{[CLS]} token to initialize the hidden state of the decoder. We reduce the dimension of the output of the decoder by half via a learned linear projection before the output softmax layer.

\section{Multitask Learning Methods}\label{ax:mtl}

Our multitask learning experiments have three somewhat distinctive properties: (i) We mix tasks with very different amounts of training data---at the extreme, under 1,000 examples for WNLI, and over 1,000,000,000 examples from LM BWB. (ii) Our goal is to optimize the quality of the shared encoder, not the performance of any one of the tasks in the multitask mix. (iii) We mix a relatively large number of tasks, up to eighteen at once in some conditions. These conditions make it challenging but important to avoid overfitting or underfitting any of our tasks.

Relatively little work has been done on this problem, so we conduct a small experiment here. All our experiments use the basic paradigm of randomly sampling a new task to train on at each step, and we experiment with two hyperparameters that can be used to control over- and underfitting: The probability with which we sample each task and the weight with which we scale the loss for each task. Our experiments follow the setup in Appendix~\ref{ax:hyper}, and do not use the ELMo BiLSTM.
For validation metrics like perplexity that decrease from high starting values during training, we include the transformed metric $1 - \frac{metric}{250}$ in our average, where the constant 250 was tuned in early experiments.

\paragraph{Task Sampling}
We consider several approaches to determine the probability with which to sample a task during training, generally making this probability a function of the amount of data available for the task. For task $i$ with training set size $N_i$, the probability is $p_i = {f(N_i)}/\sum_j{f(N_j)}$, where $f(N_i) = 1$ (Uniform), $N_i$ (Proportional), $log(N_i)$ (Log Proportional), or $N_i^{a}$ (Power $a$) where $a$ is a constant.

\begin{table}[t]
\centering \fontsize{8.4}{10.1}\selectfont \setlength{\tabcolsep}{0.5em}
\begin{tabular}{lrrrr}
\toprule
& \multicolumn{4}{c}{\textbf{Pretraining Tasks}} \\ 
\cmidrule{2-5}
\textbf{Sampling} & \textbf{GLUE} & \textbf{S1} & \textbf{S2} & \textbf{S3} \\
\midrule
\textbf{Uniform} & 69.1 & 53.7 & 82.1 & 31.7 \\ 
\textbf{Proportional} & \bf 69.8 & 52.0 & 83.1 & 36.6 \\ 
\textbf{Log Proportional} & 68.8 & 54.3 & 82.9 & 31.2 \\ 
\textbf{Power 0.75} & 69.3 & 51.1 & 82.7 & \bf 37.9 \\ 
\textbf{Power 0.66} & 69.0 & 53.4 & 82.8 & 35.5 \\ 
\textbf{Power 0.5} & 69.1 & \bf 55.6 & \bf 83.3 & 35.9 \\ 
\bottomrule
\end{tabular}
\caption{Comparison of sampling methods on four subsets of GLUE using uniform loss scaling. The reported scores are averages of the development set results achieved for each task after early stopping. Results in \textbf{bold} are the best within each set.}
\label{tab:sampling} 
\end{table}

\begin{table}[t]
\centering \fontsize{8.4}{10.1}\selectfont \setlength{\tabcolsep}{0.5em}
\begin{tabular}{lrrr}
\toprule
& \multicolumn{3}{c}{\textbf {Loss Scaling}} \\ 
\cmidrule{2-4}
\textbf {Sampling} & \textbf{Uniform} & \textbf{Proportional} & \textbf{Power 0.75} \\
\midrule
\textbf{Uniform} & 69.1 & \bf 69.7 & \bf 69.8 \\
\textbf{Proportional} & \bf 69.8 & 69.4 & 69.6 \\
\textbf{Log Proportional} & 68.8 & 68.9 & 68.9 \\
\textbf{Power 0.75} & 69.3 & 69.1 & 69.0 \\
\bottomrule
\end{tabular}
\caption{Combinations of sampling and loss scaling methods on GLUE tasks. Results in \textbf{bold} are tied for best overall GLUE score.}
\label{tab:sampling and scaling} 
\end{table}

\paragraph{Loss Scaling}
At each update, we scale the loss of a task with weight $w_i = f(N_i)/ max_{j}f(N_j) $, where $f(N_i) = 1$ (Uniform), $N_j$ (Proportional), or $N_j^{a}$ (Power $a$). 

\paragraph{Experiments}
For task sampling, we run experiments with multitask learning on the full set of nine GLUE tasks, as well as three subsets: single sentence tasks (S1: SST, CoLA), similarity and paraphrase tasks (S2: MRPC, STS, QQP), and inference tasks (S3: WNLI, QNLI, MNLI, RTE). The results are shown in Table~\ref{tab:sampling}.

We also experiment with several \textit{combinations} of task sampling and loss scaling methods, using only the full set of GLUE tasks. The results are shown in Table~\ref{tab:sampling and scaling}.

While no combination of methods consistently offers dramatically better performance than any other, we observe that it is generally better to apply only one of non-uniform sampling and non-uniform loss scaling  at a time rather than apply both simultaneously, as they provide roughly the same effect. Following encouraging results from earlier pilot experiments, we use power 0.75 task sampling and uniform loss scaling in the multitask learning experiments shown in Table~\ref{tab:glueresults}.

\section{Additional Target Task Correlations}\label{ax:correlations}

Tables~\ref{tab:corr-pretrain}, \ref{tab:corr-elmo}, and \ref{tab:corr-bert} respectively show the full target task correlations computed over pretraining, intermediate ELMo, and intermediate BERT experiments.

See Section~\ref{sec:analysis} for a discussion about correlations for the pretraining experiments.
The general trends in correlation vary significantly between the three experimental settings, which we take to roughly indicate the different types of knowledge encoded in ELMo and BERT. 
The exception is that WNLI is consistently negatively correlated with the other target tasks and often the overall GLUE score. 

For intermediate ELMo experiments, correlations are generally low, with the exception of MNLI with other tasks. CoLA is negatively correlated with most other tasks, while QQP and SST are positively correlated with most tasks.

For intermediate BERT experiments, correlations with the GLUE score are quite high, as we found that intermediate training often negatively impacted GLUE score.
QQP is highly negatively correlated with most other tasks, while the smaller tasks like MRPC and RTE are most highly correlated with overall GLUE score.

\section{Additional Learning Curves}\label{ax:curves}

\begin{figure*}[t]
\centering
\includegraphics[width=0.95\linewidth]{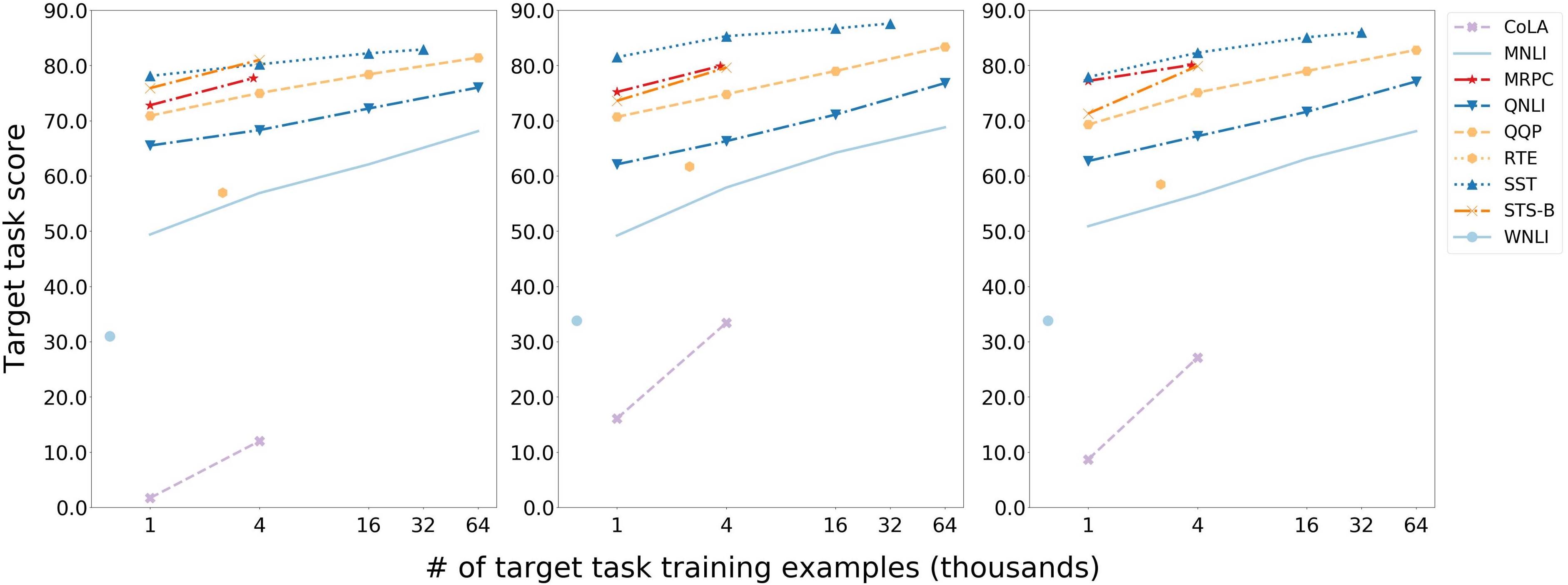}
\caption{Target-task training learning curves for each GLUE task with three encoders: the random encoder without ELMo (left), random with ELMo (center), and MTL Non-GLUE pretraining (right).
\label{fig:ax-lc}}
\end{figure*}

Figure \ref{fig:ax-lc} shows learning curves reflecting the amount of target-task data required to train a model on each GLUE task, starting from three selected encoders.
See Section~\ref{sec:analysis} for discussion.

\section{Diagnostic Set Results}\label{ax:diag}

Tables~\ref{tab:diagnostics-pretrain} and \ref{tab:diagnostics-intermediate} show results on the four coarse-grained categories of the GLUE diagnostic set for all our pretraining experiments. This set consists of about 1000 expert-constructed examples in NLI format meant to isolate a range of relevant phenomena. Results use the target task classifier trained on the MNLI training set. 

No model achieves performance anywhere close to human-level performance, suggesting that \textit{either} none of our pretrained models extract features that are suitable for robust reasoning over text, or that the MNLI training set and the MNLI target-task model are not able to exploit any such features that exist. See Section~\ref{sec:analysis} for further discussion.

While no model achieves near-human performance, the use of ELMo and appears to be helpful on examples that highlight world knowledge and lexical-semantic knowledge, and less so on examples that highlight complex logical reasoning patterns or alternations in sentence structure. This relative weakness on sentence structure is somewhat surprising given the finding in \citet{zhang2018lessons} that language model pretraining is helpful for tasks involving sentence structure.

Using BERT helps significantly with understanding sentence structure, lexical knowledge, and logical reasoning, but does not seem to help on world knowledge over using ELMo. Encouragingly, we find that intermediate training of BERT on all of our pretraining tasks outperforms intermediate training on one or no tasks in two of the four categories.

\begin{table*}[t]
\centering \fontsize{8.4}{10.1}\selectfont \setlength{\tabcolsep}{0.5em}
\input{tables/correlations-pretrain.tex}
\caption{\label{tab:corr-pretrain}Pearson correlations between performances on different target tasks, measured over all pretraining runs reported in Table \ref{tab:glueresults}.}
\end{table*}

\begin{table*}[t]
\centering \fontsize{8.4}{10.1}\selectfont \setlength{\tabcolsep}{0.5em}
\input{tables/correlations-elmo.tex}
\caption{\label{tab:corr-elmo}Pearson correlations between performances on different target tasks, measured over all ELMo runs reported in Table \ref{tab:intermediate}.
Negative correlations are \underline{underlined}.}
\end{table*}

\begin{table*}[t]
\centering \fontsize{8.4}{10.1}\selectfont \setlength{\tabcolsep}{0.5em}
\input{tables/correlations-bert.tex}
\caption{\label{tab:corr-bert}Pearson correlations between performances on different target tasks, measured over all BERT runs reported in Table \ref{tab:intermediate}.
Negative correlations are \underline{underlined}.}
\end{table*}

\begin{table*}[h]
\centering \fontsize{8.4}{10.1}\selectfont \setlength{\tabcolsep}{0.5em}
\input{tables/diagnostics-pretrain.tex}
\caption{GLUE diagnostic set results, reported as $R_3$ correlation coefficients ($\times100$), which standardizes the score of random guessing by an uninformed model at roughly 0. Human performance on the overall diagnostic set is roughly 80. Results in \textbf{bold} are the best overall.}
\label{tab:diagnostics-pretrain}
\end{table*}

\begin{table*}[h]
\centering \fontsize{8.4}{10.1}\selectfont \setlength{\tabcolsep}{0.5em}
\input{tables/diagnostics-intermediate.tex}
\caption{GLUE diagnostic set results, reported as $R_3$ correlation coefficients ($\times100$), which standardizes the score of random guessing by an uninformed model at roughly 0. Human performance on the overall diagnostic set is roughly 80. Results in \textbf{bold} are the best by section.}
\label{tab:diagnostics-intermediate}
\end{table*}

\end{document}

%% file: tables/tasks.tex
\begin{tabular}{llll}
\toprule
Task & $|$Train$|$ & Task Type \\
\midrule
\multicolumn{4}{c}{GLUE Tasks} \\
\midrule
CoLA & 8.5K & acceptability \\
SST & 67K & sentiment \\
MRPC & 3.7K & paraphrase detection \\
QQP & 364K & paraphrase detection \\
STS & 7K & sentence similarity \\
MNLI & 393K & NLI \\
QNLI & 105K & QA (NLI) \\
RTE & 2.5K & NLI \\
WNLI & 634 & coreference resolution (NLI) \\
\midrule
\multicolumn{4}{c}{Outside Tasks} \\
\midrule
DisSent WT & 311K & discourse marker prediction \\
LM WT & 4M & language modeling \\
LM BWB & 30M & language modeling \\
MT En-De & 3.4M & translation \\
MT En-Ru & 3.2M & translation \\
Reddit & 18M & response prediction \\
SkipThought & 4M & next sentence prediction & \\
\bottomrule
\end{tabular}

%% file: tables/pretraining.tex
\begin{tabular}{lr@{\hskip .1in}rrr@{/}rr@{/}rr@{/}rrrrr}
\toprule
\textbf{Pretr.} & \multicolumn{1}{c}{\textbf{Avg}} & \multicolumn{1}{c}{\textbf{CoLA}} & \multicolumn{1}{c}{\textbf{SST}} & \multicolumn{2}{c}{\textbf{MRPC}} & \multicolumn{2}{c}{\textbf{QQP}} & \multicolumn{2}{c}{\textbf{STS}} & \multicolumn{1}{c}{\textbf{MNLI}} & \multicolumn{1}{c}{\textbf{QNLI}} & \multicolumn{1}{c}{\textbf{RTE}} & \multicolumn{1}{c}{\textbf{WNLI}} \\
\midrule
\midrule
\multicolumn{14}{c}{Baselines} \\ 
\midrule
\textbf{Random} & 68.2 & 16.9 & 84.3 & 77.7 & 85.6 & 83.0 & 80.6& 81.7 & 82.6 & 73.9 & \textbf{79.6} & 57.0 & 31.0* \\
\textbf{Single-Task} & 69.1 & 21.3 & 89.0 & 77.2 & 84.7 & 84.7 & 81.9& 81.4 & 82.2& 74.8 & 78.8 & 56.0 & 11.3* \\
\midrule
\multicolumn{14}{c}{GLUE Tasks as Pretraining Tasks} \\
\midrule
\textbf{CoLA}& 68.2 & 21.3 & 85.7 & 75.0 & 83.7 & 85.7 & 82.4& 79.0 & 80.3& 72.7 & 78.4 & 56.3 & 15.5* \\
\textbf{SST} & \underline{68.6} & 16.4 & 89.0 & 76.0 & 84.2 & 84.4 & 81.6& 80.6 & 81.4& 73.9 & 78.5 & 58.8 & 19.7* \\
\textbf{MRPC}& 68.2 & 16.4 & 85.6 & 77.2 & 84.7 & 84.4 & 81.8& 81.2 & 82.2& 73.6 & 79.3 & 56.7 & 22.5* \\
\textbf{QQP} & 68.0 & 14.7 & 86.1 & 77.2 & 84.5 & 84.7 & 81.9& 81.1 & 82.0& 73.7 & 78.2 & 57.0 & 45.1* \\
\textbf{STS} & 67.7 & 14.1 & 84.6 & 77.9 & 85.3 & 81.7 & 79.2& 81.4 & 82.2& 73.6 & 79.3 & 57.4 & 43.7* \\
\textbf{MNLI}& \underline{69.1} & 16.7 & 88.2 & 78.9 & 85.2 & 84.5 & 81.5& 81.8 & 82.6& 74.8 & \textbf{79.6} & 58.8 & 36.6* \\
\textbf{QNLI}& 67.9 & 15.6 & 84.2 & 76.5 & 84.2 & 84.3 & 81.4& 80.6 & 81.8& 73.4 & 78.8 & 58.8 & \textbf{56.3}\hphantom{*} \\
\textbf{RTE} & 68.1 & 18.1 & 83.9 & 77.5 & 85.4 & 83.9 & 81.2& 81.2 & 82.2& 74.1 & 79.1 & 56.0 & 39.4* \\
\textbf{WNLI}& 68.0 & 16.3 & 84.3 & 76.5 & 84.6 & 83.0 & 80.5& 81.6 & 82.5& 73.6 & 78.8 & 58.1 & 11.3* \\

\midrule
\multicolumn{14}{c}{Non-GLUE Pretraining Tasks} \\
\midrule
\textbf{DisSent WT} & \underline{68.6} & 18.3 & 86.6 & 79.9 & 86.0 & 85.3 & 82.0& 79.5 & 80.5& 73.4 & 79.1 & 56.7 & 42.3* \\
\textbf{LM WT} & \underline{70.1} & 30.8 & 85.7 & 76.2 & 84.2 & 86.2 & 82.9& 79.2 & 80.2& 74.0 & 79.4 & 60.3 & 25.4* \\
\textbf{LM BWB} & \underline{\textbf{70.4}} & 30.7 & 86.8 & 79.9 & 86.2 & \textbf{86.3} & \textbf{83.2} & 80.7 & 81.4& 74.2 & 79.0 & 57.4 & 47.9* \\
\textbf{MT En-De}& 68.1 & 16.7 & 85.4 & 77.9 & 84.9 & 83.8 & 80.5& 82.4 & 82.9& 73.5 & \textbf{79.6} & 55.6 & 22.5* \\
\textbf{MT En-Ru}& \underline{68.4} & 16.8 & 85.1 & 79.4 & 86.2 & 84.1 & 81.2& 82.7 & 83.2& 74.1 & 79.1 & 56.0 & 26.8* \\
\textbf{Reddit} & 66.9 & 15.3 & 82.3 & 76.5 & 84.6 & 81.9 & 79.2& 81.5 & 81.9& 72.7 & 76.8 & 55.6 & 53.5* \\
\textbf{SkipThought} & \underline{68.7} & 16.0 & 84.9 & 77.5 & 85.0 & 83.5 & 80.7& 81.1 & 81.5 & 73.3 & 79.1 & \textbf{63.9} & 49.3* \\

\midrule
\multicolumn{14}{c}{Multitask Pretraining} \\
\midrule
\textbf{MTL GLUE}& \underline{68.9} & 15.4 & \textbf{89.9} & 78.9 & 86.3 & 82.6 & 79.9 & \textbf{82.9} & \textbf{83.5} & \textbf{74.9}& 78.9 & 57.8 & 38.0* \\
\textbf{MTL Non-GLUE} & \underline{69.9} & 30.6 & 87.0 & \textbf{81.1} & \textbf{87.6} & 86.0 & 82.2 & 79.9 & 80.6 & 72.8 & 78.9 & 54.9 & 22.5* \\
\textbf{MTL All} & \underline{\textbf{70.4}} & \textbf{33.2} & 88.2 & 78.9 & 85.9 & 85.5 & 81.8 & 79.7 & 80.0 & 73.9 & 78.7 & 57.4 & 33.8* \\

\midrule
\midrule
\multicolumn{14}{c}{\it Test Set Results} \\
\midrule
\textbf{LM BWB} & 66.5 & 29.1 & 86.9 & 75.0 & 82.1 & 82.7 & 63.3 & 74.0 & 73.1 & 73.4 & 68.0 & 51.3 & 65.1 \\
\textbf{MTL All} & 68.5 & 36.3 & 88.9 & 77.7 & 84.8 & 82.7 & 63.6 & 77.8 & 76.7 & 75.3 & 66.2 & 53.2 & 65.1\\
\bottomrule
\end{tabular}

%% file: tables/intermediate-by-model.tex
\begin{tabular}{lr@{\hskip .1in}rrr@{/}rr@{/}rr@{/}rrrrr}
\toprule
\textbf{Intermediate Task} & \multicolumn{1}{c}{\textbf{Avg}} & \multicolumn{1}{c}{\textbf{CoLA}} & \multicolumn{1}{c}{\textbf{SST}} & \multicolumn{2}{c}{\textbf{MRPC}} & \multicolumn{2}{c}{\textbf{QQP}} & \multicolumn{2}{c}{\textbf{STS}} & \multicolumn{1}{c}{\textbf{MNLI}} & \multicolumn{1}{c}{\textbf{QNLI}} & \multicolumn{1}{c}{\textbf{RTE}} & \multicolumn{1}{c}{\textbf{WNLI}} \\
\midrule
\midrule
\multicolumn{14}{c}{ELMo with Intermediate Task Training} \\ 
\midrule
\textbf{Random}$^E$ & 70.5  & 38.5& 87.7 & 79.9  & 86.5 & 86.7 & 83.4  & 80.8  & 82.1& 75.6  & 79.6& \textbf{61.7}  & 33.8* \\
\textbf{Single-Task}$^E$ & 71.2 & 39.4& \textbf{90.6} & 77.5  & 84.4 & 86.4 & 82.4  &79.9  & 80.6  & 75.6  & 78.0& 55.6  & 11.3* \\

\textbf{CoLA}$^E$ & 71.1  & 39.4& 87.3 & 77.5  & 85.2 & 86.5 & 83.0  & 78.8  & 80.2& 74.2  & 78.2& 59.2  & 33.8* \\
\textbf{SST}$^E$ & 71.2  & 38.8& \textbf{90.6} & 80.4  & 86.8 & 87.0 & 83.5  & 79.4  & 81.0& 74.3  & 77.8& 53.8  & 43.7* \\
\textbf{MRPC}$^E$ & \underline{71.3} & 40.0& 88.4 & 77.5  & 84.4 & 86.4 & 82.7  & 79.5  & 80.6& 74.9  & 78.4& 58.1  & \textbf{54.9}* \\
\textbf{QQP}$^E$ & 70.8  & 34.3& 88.6 & 79.4  & 85.7 & 86.4 & 82.4  & 81.1  & 82.1& 74.3  & 78.1& 56.7  & 38.0* \\
\textbf{STS}$^E$ &  \underline{71.6} & 39.9	 & 88.4 &	79.9 &	86.4 & 86.7  & 	83.3  & 	  79.9  &	80.6	& 74.3 &	78.6 &	58.5 &	26.8*	  \\
\textbf{MNLI}$^E$ & \underline{72.1}  & 38.9& 89.0 & 80.9  & 86.9 & 86.1 & 82.7  & 81.3  & 82.5& 75.6  & 79.7& 58.8  & 16.9* \\
\textbf{QNLI}$^E$ & 71.2  & 37.2 & 88.3 & 81.1  & 86.9 & 85.5 & 81.7  & 78.9  & 80.1& 74.7  & 78.0& 58.8  & 22.5* \\
\textbf{RTE}$^E$ & 71.2  & 38.5& 87.7 & 81.1  & 87.3 & 86.6 & 83.2  & 80.1  & 81.1& 74.6  & 78.0& 55.6  & 32.4* \\
\textbf{WNLI}$^E$ & 70.9  & 38.4& 88.6 & 78.4  & 85.9 & 86.3 & 82.8  & 79.1  & 80.0& 73.9  & 77.9& 57.0  & 11.3* \\
\textbf{DisSent WT}$^E$ & \underline{71.9}  & 39.9& 87.6 & \textbf{81.9}  & \textbf{87.2} & 85.8 & 82.3  & 79.0  & 80.7& 74.6  & 79.1& 61.4  & 23.9* \\
\textbf{MT En-De}$^E$ & \underline{72.1}  & 40.1& 87.8 & 79.9  & 86.6 & 86.4 & 83.2  & 81.8  & 82.4& 75.9  & 79.4& 58.8  & 31.0* \\
\textbf{MT En-Ru}$^E$ & 70.4  & \textbf{41.0} & 86.8 & 76.5  & 85.0 & 82.5 & 76.3  & 81.4  & 81.5& 70.1  & 77.3& 60.3  & 45.1* \\
\textbf{Reddit}$^E$ & 71.0  & 38.5& 87.7 & 77.2  & 85.0 & 85.4 & 82.1  & 80.9  & 81.7& 74.2  & 79.3& 56.7  & 21.1* \\
\textbf{SkipThought}$^E$ & \underline{71.7} & 40.6& 87.7 & 79.7  & 86.5 & 85.2 & 82.1  & 81.0  & 81.7& 75.0  & 79.1& 58.1  & 52.1* \\
\textbf{MTL GLUE}$^E$ & \underline{72.1}  & 33.8& 90.5 & 81.1  & 87.4 & 86.6 & 83.0  & 82.1  & 83.3& \textbf{76.2} & 79.2& 61.4  & 42.3* \\
\textbf{MTL Non-GLUE}$^E$ & \underline{\textbf{72.4}} & 39.4& 88.8 & 80.6  & 86.8 & \textbf{87.1} & \textbf{84.1}  & \textbf{83.2} & \textbf{83.9} & 75.9  & \textbf{80.9} & 57.8  & 22.5* \\
\textbf{MTL All$^E$} & \underline{72.2} & 37.9 & 89.6 & 79.2 & 86.4 & 86.0 & 82.8 & 81.6 & 82.5 & 76.1 & 80.2 & 60.3 & 31.0* \\

\midrule
\multicolumn{14}{c}{BERT with Intermediate Task Training} \\
\midrule
\textbf{Single-Task}$^B$ & 78.8 & 56.6 & 90.9 & 88.5 & 91.8 & 89.9 & 86.4 & 86.1 & 86.0 & 83.5 & \textbf{87.9} & 69.7 & \textbf{56.3}\hphantom{*} \\
\textbf{CoLA}$^B$ & 78.3 & \textbf{61.3} & 91.1 & 87.7 & 91.4 & 89.7 & 86.3 & 85.0 & 85.0 & 83.3 & 85.9 & 64.3 & 43.7* \\
\textbf{SST}$^B$ & 78.4 & 57.4 & \textbf{92.2} & 86.3 & 90.0 & 89.6 & 86.1 & 85.3 & 85.1 & 83.2 & 87.4 & 67.5 & 43.7* \\
\textbf{MRPC}$^B$ & 78.3 & 60.3 & 90.8 & 87.0 & 91.1 & 89.7 & 86.3 & 86.6 & 86.4 & \textbf{83.8} & 83.9 & 66.4 & \textbf{56.3}\hphantom{*} \\
\textbf{QQP}$^B$ & \underline{79.1} & 56.8 & 91.3 & 88.5 & 91.7 & \textbf{90.5} & \textbf{87.3} & 88.1 & 87.8 & 83.4 & 87.2 & 69.7 & \textbf{56.3}\hphantom{*} \\
\textbf{STS}$^B$ & \underline{79.4} & 61.1 & 92.3 & 88.0 & 91.5 & 89.3 & 85.5 & 86.2 & 86.0 & 82.9 & 87.0 & 71.5 & 50.7* \\
\textbf{MNLI}$^B$ & \underline{\textbf{79.6}} & 56.0 & 91.3 & 88.0 & 91.3 & 90.0 & 86.7 & 87.8 & 87.7 & 82.9 & 87.0 & \textbf{76.9} & \textbf{56.3}\hphantom{*} \\
\textbf{QNLI}$^B$ & 78.4 & 55.4 & 91.2 & \textbf{88.7} & \textbf{92.1} & 89.9 & 86.4 & 86.5 & 86.3 & 82.9 & 86.8 & 68.2 & \textbf{56.3}\hphantom{*} \\
\textbf{RTE}$^B$ & 77.7 & 59.3 & 91.2 & 86.0 & 90.4 & 89.2 & 85.9 & 85.9 & 85.7 & 82.0 & 83.3 & 65.3 & \textbf{56.3}\hphantom{*} \\
\textbf{WNLI}$^B$ & 76.2 & 53.2 & 92.1 & 85.5 & 90.0 & 89.1 & 85.5 & 85.6 & 85.4 & 82.4 & 82.5 & 58.5 & \textbf{56.3}\hphantom{*} \\
\textbf{DisSent WT}$^B$ & 78.1 & 58.1 & 91.9 & 87.7 & 91.2 & 89.2 & 85.9 & 84.2 & 84.1 & 82.5 & 85.5 & 67.5 & 43.7* \\
\textbf{MT En-De}$^B$ & 73.9 & 47.0 & 90.5 & 75.0 & 83.4 & 89.6 & 86.1 & 84.1 & 83.9 & 81.8 & 83.8 & 54.9 & \textbf{56.3}\hphantom{*} \\
\textbf{MT En-Ru}$^B$ & 74.3 & 52.4 & 89.9 & 71.8 & 81.3 & 89.4 & 85.6 & 82.8 & 82.8 & 81.5 & 83.1 & 58.5 & 43.7* \\
\textbf{Reddit}$^B$ & 75.6 & 49.5 & 91.7 & 84.6 & 89.2 & 89.4 & 85.8 & 83.8 & 83.6 & 81.8 & 84.4 & 58.1 & \textbf{56.3}\hphantom{*} \\
\textbf{SkipThought}$^B$ & 75.2 & 53.9 & 90.8 & 78.7 & 85.2 & 89.7 & 86.3 & 81.2 & 81.5 & 82.2 & 84.6 & 57.4 & 43.7* \\
\textbf{MTL GLUE}$^B$ & \underline{\textbf{79.6}} & 56.8 & 91.3 & 88.0 & 91.4 & 90.3 & 86.9 & \textbf{89.2} & \textbf{89.0} & 83.0 & 86.8 & 74.7 & 43.7* \\
\textbf{MTL Non-GLUE}$^B$ & 76.7 & 54.8 & 91.1 & 83.6 & 88.7 & 89.2 & 85.6 & 83.2 & 83.2 & 82.4 & 84.4 & 64.3 & 43.7* \\
\textbf{MTL All}$^B$ & \underline{79.3} & 53.1 & 91.7 & 88.0 & 91.3 & 90.4 & 87.0 & 88.1 & 87.9 & 83.5 & 87.6 & 75.1 & 45.1* \\

\midrule
\midrule
\multicolumn{14}{c}{\it Test Set Results} \\
\midrule
\textbf{Non-GLUE$^E$} & 69.7 & 34.5 & 89.5 & 78.2 & 84.8 & 83.6 & 64.3 & 77.5 & 76.0 & 75.4 & 74.8 & 55.6 & 65.1\\
\textbf{MNLI$^B$} & 77.1 & 49.6 & 93.2 & 88.5 & 84.7 & 70.6 & 88.3 & 86.0 & 85.5 & 82.7 & 78.7 & 72.6 & 65.1 \\
\textbf{GLUE$^B$} & 77.3 & 49.0 & 93.5 & 89.0 & 85.3 & 70.6 & 88.6 & 85.8 & 84.9 & 82.9 & 81.0 & 71.7 & 34.9 \\
\textbf{BERT Base} & 78.4 & 52.1 & 93.5 & 88.9 & 84.8 & 71.2 & 89.2 & 87.1 & 85.8 & 84.0 & 91.1 & 66.4 & 65.1 \\
\bottomrule
\end{tabular}

%% file: tables/correlations-pretrain-subset.tex
\begin{tabular}{lrrrrrrrr}
\toprule
\textbf{Task} & \textbf{Avg}   & \textbf{CoLA}  & \textbf{SST} & \textbf{STS} & \textbf{QQP} & \textbf{MNLI} & \textbf{QNLI} \\
\midrule
\bf CoLA & 0.86 & 1.00 &  &  &  &  &  \\
\bf SST  & 0.60 & 0.25 & 1.00 &  &  &  &   \\
\bf MRPC & 0.39 & 0.21 & 0.34 &  &  &  &  \\
\bf STS  & \underline{-0.36} & \underline{-0.60} & 0.01 & 1.00 &  &  &  \\
\bf QQP & 0.61 & 0.61 & 0.27 & \underline{-0.58} & 1.00 &  &  \\
\bf MNLI & 0.54 & 0.16 & 0.66 & 0.40 & 0.08 & 1.00 &  \\
\bf QNLI & 0.43 & 0.13 & 0.26 & 0.04 & 0.27 & 0.56 & 1.00 \\
\bf RTE  & 0.34 & 0.08 & 0.16 & \underline{-0.10} & 0.04 & 0.14 & 0.32 \\
\bf WNLI & \underline{-0.21} & \underline{-0.21} & \underline{-0.37} & 0.31 & \underline{-0.37} & \underline{-0.07} & \underline{-0.26} \\
\bottomrule
\end{tabular}

%% file: tables/correlations-pretrain.tex
\begin{tabular}{lrrrrrrrrrr}
\toprule
\textbf{Task} & \textbf{Avg}   & \textbf{CoLA}  & \textbf{SST} & \textbf{MRPC} & \textbf{STS} & \textbf{QQP} & \textbf{MNLI} & \textbf{QNLI}  & \textbf{RTE}   & \textbf{WNLI} \\
\midrule
\bf CoLA & 0.86 & 1.00 &  &  &  &  &  &  &  &  \\
\bf SST  & 0.60 & 0.25 & 1.00 &  &  &  &  &  &  &  \\
\bf MRPC & 0.39 & 0.21 & 0.34 & 1.00 &  &  &  &  &  & \\
\bf STS  & \underline{-0.36} & \underline{-0.60} & 0.01 & 0.29 & 1.00 &  &  &  &  & \\
\bf QQP  & 0.61 & 0.61 & 0.27 & \underline{-0.17} & \underline{-0.58} & 1.00 &  &  &  & \\
\bf MNLI & 0.54 & 0.16 & 0.66 & 0.56 & 0.40 & 0.08 & 1.00 &  &  & \\
\bf QNLI & 0.43 & 0.13 & 0.26 & 0.32 & 0.04 & 0.27 & 0.56 & 1.00 &  & \\
\bf RTE  & 0.34 & 0.08 & 0.16 & \underline{-0.09} & \underline{-0.10} & 0.04 & 0.14 & 0.32 & 1.00 & \\
\bf WNLI & \underline{-0.21} & \underline{-0.21} & \underline{-0.37} & 0.31 & 0.31 & \underline{-0.37} & \underline{-0.07} & \underline{-0.26} & 0.12 & 1.00 \\
\bottomrule
\end{tabular}

%% file: tables/correlations-elmo.tex
\begin{tabular}{lrrrrrrrrrr}
\toprule
\textbf{Task} & \textbf{Avg}   & \textbf{CoLA}  & \textbf{SST} & \textbf{MRPC} & \textbf{STS} & \textbf{QQP} & \textbf{MNLI} & \textbf{QNLI}  & \textbf{RTE}   & \textbf{WNLI} \\
\midrule
\bf CoLA & 0.07 & 1.00 &  &  &  &  &  &  &  & \\
\bf SST  & 0.32 & \underline{-0.48} & 1.00 &  &  &  &  &  &  & \\
\bf MRPC & 0.42 & \underline{-0.20} & 0.29 & 1.00 &  &  &  &  &  & \\
\bf STS & 0.41 & \underline{-0.40} & 0.26 & 0.21 & 1.00 &  &  &  &  & \\
\bf QQP & 0.02 & 0.08 & 0.26 & 0.18 & 0.15 & 1.00 &  &  &  & \\
\bf MNLI & 0.60 & \underline{-0.21} & 0.33 & 0.38 & 0.72 & 0.21 & 1.00 &  &  & \\
\bf QNLI & 0.50 & 0.10 & 0.03 & 0.12 & 0.63 & \underline{-0.01} & 0.72 & 1.00 &  & \\
\bf RTE &  0.39 & \underline{-0.13} & \underline{-0.15} & 0.21 & 0.27 & \underline{-0.04} & 0.60 & 0.59 & 1.00 & \\
\bf WNLI & \underline{-0.14} & 0.02 & 0.23 & \underline{-0.29} & \underline{-0.02} & 0.15 & 0.02 & \underline{-0.25} & \underline{-0.22} & 1.00 \\
\bottomrule
\end{tabular}

%% file: tables/correlations-bert.tex
\begin{tabular}{lrrrrrrrrrr}
\toprule
\textbf{Task} & \textbf{Avg}   & \textbf{CoLA}  & \textbf{SST} & \textbf{MRPC} & \textbf{STS} & \textbf{QQP} & \textbf{MNLI} & \textbf{QNLI}  & \textbf{RTE}   & \textbf{WNLI} \\
\midrule
\bf CoLA & 0.71 & 1.00 &  &  &  &  &  &  &  & \\
\bf SST  & 0.41 & 0.32 & 1.00 &  &  &  &  &  &  &  \\
\bf MRPC & 0.83 & 0.67 & 0.62 & 1.00 &  &  &  &  &  & \\
\bf STS & 0.82 & 0.34 & 0.21 & 0.60 & 1.00 &  &  &  &  & \\
\bf QQP & \underline{-0.41} & 0.01 & 0.04 & \underline{-0.05} & \underline{-0.64} & 1.00 &  &  &  &  \\
\bf MNLI & 0.73 & 0.31 & 0.10 & 0.42 & 0.69 & \underline{-0.68} & 1.00 &  &  & \\
\bf QNLI & 0.73 & 0.38 & 0.29 & 0.56 & 0.43 & \underline{-0.11} & 0.62 & 1.00 &  &  \\
\bf RTE & 0.88 & 0.47 & 0.22 & 0.56 & 0.87 & \underline{-0.70} & 0.68 & 0.55 & 1.00 & \\
\bf WNLI & 0.45 & \underline{-0.10} & \underline{-0.03} & 0.20 & 0.79 & \underline{-0.89} & 0.65 & 0.11 & 0.69 & 1.00 \\
\bottomrule
\end{tabular}

%% file: tables/diagnostics-pretrain.tex
\begin{tabular}{lrrrr}
\toprule
\textbf{Pretr.} & \multicolumn{1}{c}{\textbf{Knowledge}} &  \multicolumn{1}{c}{\textbf{Lexical Semantics}} & \multicolumn{1}{c}{\textbf{Logic}} &  \multicolumn{1}{c}{\textbf{Predicate/Argument Str.}} \\
\midrule
\midrule
\multicolumn{5}{c}{Baselines} \\
\midrule
\textbf{Random}                      & 17.6                   & 19.6                & 12.5               & 26.9                 \\
\midrule
\multicolumn{5}{c}{GLUE Tasks as Pretraining Tasks} \\
\midrule
\textbf{CoLA}                     & 15.3                   & 24.2                & 14.9               & \bf \textbf{31.7}                 \\
\textbf{SST}                      & 16.1                   & 24.8                & 16.5               & 28.7                 \\
\textbf{MRPC}                     & 16.0                   & \bf \textbf{25.2}                & 12.6               & 26.4                 \\
\textbf{QQP}                      & 12.8                   & 22.5                & 12.9               & 30.8                 \\
\textbf{STS}                    & 16.5                   & 20.2                & 13.0               & 27.1                 \\
\textbf{MNLI}                     & 16.4                   & 20.4                & \bf \textbf{17.7}               & 29.9                 \\
\textbf{QNLI}                     & 13.6                   & 21.3                & 12.2               & 28.0                 \\
\textbf{RTE}                      & 16.3                   & 23.1                & 14.5               & 28.8                 \\
\textbf{WNLI}                     & \textbf{18.8}                   & 19.5                & 13.9               & 29.1                 \\
\midrule
\multicolumn{5}{c}{Non-GLUE Pretraining Tasks} \\
\midrule
\textbf{DisSent WP}                  & 18.5                   & 24.2                & 15.4               & 27.8                 \\
\textbf{LM WP}                   & 14.9                   & 16.6                & 9.4                & 23.0                 \\
\textbf{LM BWB}                   & 15.8                   & 19.4                & 9.1                & 23.9                 \\
\textbf{MT En-De}                    & 13.4                   & 24.6                & 14.8               & 30.1                 \\
\textbf{MT En-Ru}                    & 13.4                   & 24.6                & 14.8               & 30.1                 \\
\textbf{Reddit}               & 13.9                   & 20.4                & 14.1               & 26.0                 \\
\textbf{SkipThought}                   & 15.1                   & 22.0                & 13.7               & 27.9                 \\
\midrule
\multicolumn{5}{c}{Multitask Pretraining} \\
\midrule
\textbf{MTL All}                      & 16.3 & 21.4 & 11.2 & 28.0           \\
\textbf{MTL GLUE}                     & 12.5 &	21.4 &	15.0 &	30.1            \\
\textbf{MTL Outside}                  & 14.5&	19.7&	13.1&	26.2              \\
\bottomrule
\end{tabular}

%% file: tables/diagnostics-intermediate.tex
\begin{tabular}{lrrrr}
\toprule
\textbf{Pretr.} & \multicolumn{1}{c}{\textbf{Knowledge}} &  \multicolumn{1}{c}{\textbf{Lexical Semantics}} & \multicolumn{1}{c}{\textbf{Logic}} &  \multicolumn{1}{c}{\textbf{Predicate/Argument Str.}} \\
\midrule
\midrule
\multicolumn{5}{c}{ELMo with Intermediate Task Training} \\
\midrule
\textbf{Random}$^E$             & 19.2                   & 22.9                & 9.8                & 25.5                 \\
\textbf{CoLA}$^E$             & 17.2                   & 21.6                & 9.2                & 27.3                 \\
\textbf{SST}$^E$             & 19.4                   & 20.5                & 9.7                & 28.5                 \\
\textbf{MRPC}$^E$             & 11.8                   & 20.5                & 12.1               & 27.4                 \\
\textbf{QQP}$^E$             & 17.5                   & 16.0                & 9.9                & \textbf{30.5}                 \\
\textbf{STS}$^E$            & 18.0                         &  18.4                   & 9.1                   &   25.5                   \\
\textbf{MNLI}$^E$             & 17.0                   & 23.2                & 14.4               & 23.9                 \\
\textbf{QNLI}$^E$             & 17.4                   & \textbf{24.1}                & 10.7               & 30.2                 \\
\textbf{RTE}$^E$             & 18.0                   & 20.2                & 8.7                & 28.0                 \\
\textbf{WNLI}$^E$             & 16.5                   & 19.8                & 7.3                & 25.2                 \\

\textbf{DisSent WP}$^E$             & 16.3                   & 23.0                & 11.6               & 26.5                 \\
\textbf{MT En-De}$^E$             & 19.2                   & 21.0                & 13.5               & 29.7                 \\
\textbf{MT En-Ru}$^E$             & 20.0                   & 20.1                & 11.9               & 21.4                 \\
\textbf{Reddit}$^E$             & 14.7                   & 22.3                & \textbf{15.0}               & 29.0                 \\
\textbf{SkipThought}$^E$             & 20.5                   & 18.5                & 10.4               & 26.8 \\ 
\textbf{MTL GLUE}$^E$             & \bf 20.6                   & 22.1                & 14.7               & 25.3                 \\
\textbf{MTL Non-GLUE}$^E$             & 15.7                   & 23.7                & 12.6               & 29.0                 \\
\textbf{MTL All}$^E$             &   13.8 &	18.4	 & 10.8 &	26.7  \\
\midrule
\multicolumn{5}{c}{BERT with Intermediate Task Training} \\
\midrule
\textbf{Single-Task}$^B$ & 20.3 & 36.3 & 21.7 & 40.4 \\ 
\textbf{CoLA}$^B$ &18.5 & 34.0 & 23.5 & 40.1 \\ 
\textbf{SST}$^B$  &19.8 & 36.0 & 23.2 & 39.1 \\
\textbf{MRPC}$^B$ &20.6 & 33.3 & 20.9 & 37.8 \\
\textbf{QQP}$^B$  &17.4 & 35.7 & 23.8 & 40.5 \\
\textbf{STS}$^B$  & \textbf{21.3} & 34.7 & 24.0 & 40.7 \\
\textbf{MNLI}$^B$ &19.1 & 34.0 & 23.3 & \textbf{41.7} \\
\textbf{QNLI}$^B$ &20.3 & 38.4 & 24.4 & 41.5 \\
\textbf{RTE}$^B$  &15.4 & 32.6 & 20.2 & 38.5 \\
\textbf{WNLI}$^B$ &20.8 & 35.8 & 23.1 & 39.3 \\
\textbf{DisSent WP}$^B$ & 17.9 & 34.0 & 23.7 & 39.1\\
\textbf{MT En-De}$^B$   & 18.6 & 33.8 & 20.7 & 37.4\\
\textbf{MT En-Ru}$^B$   & 14.2 & 30.2 & 20.3 & 36.5\\
\textbf{Reddit}$^B$     & 16.5 & 29.9 & 22.7 & 37.1\\
\textbf{SkipThought}$^B$& 15.8 & 35.0 & 20.9 & 38.3\\
\textbf{MTL GLUE}$^B$       & 17.0 & 35.2 & 24.3 & 39.6\\
\textbf{MTL Non-GLUE}$^B$   & 18.7 & 37.0 & 21.8 & 40.6\\
\textbf{MTL All}$^B$        & 17.8 & \textbf{40.3} & \textbf{27.5} & 41.0 \\
\bottomrule
\end{tabular}

%% file: main.bbl
\begin{thebibliography}{61}
\expandafter\ifx\csname natexlab\endcsname\relax\def\natexlab#1{#1}\fi

\bibitem[{Bingel and S{\o}gaard(2017)}]{bingel-sogaard:2017:EACLshort}
Joachim Bingel and Anders S{\o}gaard. 2017.
\newblock \href {http://www.aclweb.org/anthology/E17-2026} {Identifying
  beneficial task relations for multi-task learning in deep neural networks}.
\newblock In \emph{Proceedings of the 15th Conference of the European Chapter
  of the Association for Computational Linguistics: Volume 2, Short Papers},
  pages 164--169, Valencia, Spain. Association for Computational Linguistics.

\bibitem[{Bojar et~al.(2017)Bojar, Buck, Chatterjee, Federmann, Graham, Haddow,
  Huck, Yepes, Koehn, and Kreutzer}]{W17-4700}
Ond{\v{r}}ej Bojar, Christian Buck, Rajen Chatterjee, Christian Federmann,
  Yvette Graham, Barry Haddow, Matthias Huck, Antonio~Jimeno Yepes, Philipp
  Koehn, and Julia Kreutzer. 2017.
\newblock \href {http://aclweb.org/anthology/W17-4700} {Proceedings of the
  second conference on machine translation}.
\newblock In \emph{Proceedings of the Second Conference on Machine
  Translation}. Association for Computational Linguistics.

\bibitem[{Bojar et~al.(2014)Bojar, Buck, Federmann, Haddow, Koehn, Leveling,
  Monz, Pecina, Post, Saint-Amand, Soricut, Specia, and
  Tamchyna}]{bojar2014findings}
Ondrej Bojar, Christian Buck, Christian Federmann, Barry Haddow, Philipp Koehn,
  Johannes Leveling, Christof Monz, Pavel Pecina, Matt Post, Herve Saint-Amand,
  Radu Soricut, Lucia Specia, and Ale{\v{s}} Tamchyna. 2014.
\newblock \href {https://doi.org/10.3115/v1/W14-3302} {Findings of the 2014
  workshop on statistical machine translation}.
\newblock In \emph{Proceedings of the Ninth Workshop on Statistical Machine
  Translation}, pages 12--58. Association for Computational Linguistics.

\bibitem[{Cer et~al.(2017)Cer, Diab, Agirre, Lopez-Gazpio, and
  Specia}]{cer2017semeval}
Daniel Cer, Mona Diab, Eneko Agirre, Inigo Lopez-Gazpio, and Lucia Specia.
  2017.
\newblock \href {https://doi.org/10.18653/v1/S17-2001} {Semeval-2017 task 1:
  Semantic textual similarity multilingual and crosslingual focused
  evaluation}.
\newblock In \emph{Proceedings of the 11th International Workshop on Semantic
  Evaluation (SemEval-2017)}, pages 1--14. Association for Computational
  Linguistics.

\bibitem[{Changpinyo et~al.(2018)Changpinyo, Hu, and
  Sha}]{changpinyo-hu-sha:2018:C18-1}
Soravit Changpinyo, Hexiang Hu, and Fei Sha. 2018.
\newblock \href {http://www.aclweb.org/anthology/C18-1251} {Multi-task learning
  for sequence tagging: An empirical study}.
\newblock In \emph{Proceedings of the 27th International Conference on
  Computational Linguistics}, pages 2965--2977, Santa Fe, New Mexico, USA.
  Association for Computational Linguistics.

\bibitem[{Chelba et~al.(2013)Chelba, Mikolov, Schuster, Ge, Brants, Koehn, and
  Robinson}]{chelba2013one}
Ciprian Chelba, Tomas Mikolov, Mike Schuster, Qi~Ge, Thorsten Brants, Phillipp
  Koehn, and Tony Robinson. 2013.
\newblock One billion word benchmark for measuring progress in statistical
  language modeling.
\newblock \emph{arXiv preprint 1312.3005}.

\bibitem[{Chen and Manning(2014)}]{chen2014fast}
Danqi Chen and Christopher Manning. 2014.
\newblock \href {https://doi.org/10.3115/v1/D14-1082} {A fast and accurate
  dependency parser using neural networks}.
\newblock In \emph{Proceedings of the 2014 Conference on Empirical Methods in
  Natural Language Processing (EMNLP)}, pages 740--750. Association for
  Computational Linguistics.

\bibitem[{Collobert et~al.(2011)Collobert, Weston, Bottou, Karlen, Kavukcuoglu,
  and Kuksa}]{collobert2011natural}
Ronan Collobert, Jason Weston, L{\'e}on Bottou, Michael Karlen, Koray
  Kavukcuoglu, and Pavel Kuksa. 2011.
\newblock Natural language processing (almost) from scratch.
\newblock \emph{Journal of Machine Learning Research}, 12(Aug):2493--2537.

\bibitem[{Conneau et~al.(2017)Conneau, Kiela, Schwenk, Barrault, and
  Bordes}]{DBLP:conf/emnlp/ConneauKSBB17}
Alexis Conneau, Douwe Kiela, Holger Schwenk, Lo{\"{\i}}c Barrault, and Antoine
  Bordes. 2017.
\newblock \href {http://aclanthology.info/papers/D17-1071/d17-1071} {Supervised
  learning of universal sentence representations from natural language
  inference data}.
\newblock In \emph{Proceedings of the 2017 Conference on Empirical Methods in
  Natural Language Processing, {EMNLP} 2017, Copenhagen, Denmark, September
  9-11, 2017}, pages 681--691.

\bibitem[{Dagan et~al.(2006)Dagan, Glickman, and Magnini}]{dagan2006pascal}
Ido Dagan, Oren Glickman, and Bernardo Magnini. 2006.
\newblock The {PASCAL} recognising textual entailment challenge.
\newblock In \emph{Machine learning challenges. evaluating predictive
  uncertainty, visual object classification, and recognising tectual
  entailment}, pages 177--190. Springer.

\bibitem[{Dai and Le(2015)}]{dai2015semi}
Andrew~M. Dai and Quoc~V. Le. 2015.
\newblock \href
  {http://papers.nips.cc/paper/5949-semi-supervised-sequence-learning.pdf}
  {Semi-supervised sequence learning}.
\newblock In C.~Cortes, N.~D. Lawrence, D.~D. Lee, M.~Sugiyama, and R.~Garnett,
  editors, \emph{Advances in Neural Information Processing Systems 28}, pages
  3079--3087. Curran Associates, Inc.

\bibitem[{Devlin et~al.(2019)Devlin, Chang, Lee, and
  Toutanova}]{devlin2018bert}
Jacob Devlin, Ming-Wei Chang, Kenton Lee, and Kristina Toutanova. 2019.
\newblock {BERT}: Pre-training of deep bidirectional transformers for language
  understanding.
\newblock In \emph{Proceedings of the 2019 Conference of the North American
  Chapter of the Association for Computational Linguistics: Human Language
  Technologies, Volume 1 (Long Papers)}.

\bibitem[{Dolan and Brockett(2005)}]{dolan2005automatically}
William~B. Dolan and Chris Brockett. 2005.
\newblock \href {http://www.aclweb.org/anthology/I05-5002} {Automatically
  constructing a corpus of sentential paraphrases}.
\newblock In \emph{Proceedings of the Third International Workshop on
  Paraphrasing (IWP2005)}.

\bibitem[{Gardner et~al.(2017)Gardner, Grus, Neumann, Tafjord, Dasigi, Liu,
  Peters, Schmitz, and Zettlemoyer}]{Gardner2017AllenNLP}
Matt Gardner, Joel Grus, Mark Neumann, Oyvind Tafjord, Pradeep Dasigi,
  Nelson~F. Liu, Matthew Peters, Michael Schmitz, and Luke~S. Zettlemoyer.
  2017.
\newblock {AllenNLP}: A deep semantic natural language processing platform.
\newblock \emph{arXiv preprint 1803.07640}.

\bibitem[{Hassan et~al.(2018)Hassan, Aue, Chen, Chowdhary, Clark, Federmann,
  Huang, Junczys-Dowmunt, Lewis, Li, Liu, Liu, Luo, Menezes, Qin, Seide, Tan,
  Tian, Wu, Wu, Xia, Zhang, Zhang, and
  Zhou}]{achieving-human-parity-on-automatic-chinese-to-english-news-translation}
Hany Hassan, Anthony Aue, Chang Chen, Vishal Chowdhary, Jonathan Clark,
  Christian Federmann, Xuedong Huang, Marcin Junczys-Dowmunt, Will Lewis,
  Mu~Li, Shujie Liu, Tie-Yan Liu, Renqian Luo, Arul Menezes, Tao Qin, Frank
  Seide, Xu~Tan, Fei Tian, Lijun Wu, Shuangzhi Wu, Yingce Xia, Dongdong Zhang,
  Zhirui Zhang, and Ming Zhou. 2018.
\newblock \href
  {https://www.microsoft.com/en-us/research/publication/achieving-human-parity-on-automatic-chinese-to-english-news-translation/}
  {Achieving human parity on automatic {C}hinese to {E}nglish news
  translation}.

\bibitem[{Hill et~al.(2016)Hill, Cho, and Korhonen}]{hill2016learning}
Felix Hill, Kyunghyun Cho, and Anna Korhonen. 2016.
\newblock \href {https://doi.org/10.18653/v1/N16-1162} {Learning distributed
  representations of sentences from unlabelled data}.
\newblock In \emph{Proceedings of the 2016 Conference of the North American
  Chapter of the Association for Computational Linguistics: Human Language
  Technologies}, pages 1367--1377. Association for Computational Linguistics.

\bibitem[{Hochreiter and Schmidhuber(1997)}]{hochreiter1997long}
Sepp Hochreiter and J{\"u}rgen Schmidhuber. 1997.
\newblock Long short-term memory.
\newblock \emph{Neural computation}, 9(8):1735--1780.

\bibitem[{Hockenmaier and Steedman(2007)}]{CCG}
Julia Hockenmaier and Mark Steedman. 2007.
\newblock \href {http://www.aclweb.org/anthology/J07-3004} {{CCGbank: A Corpus
  of CCG Derivations and Dependency Structures Extracted from the Penn
  Treebank}}.
\newblock \emph{Computational Linguistics}.

\bibitem[{Houlsby et~al.(2019)Houlsby, Giurgiu, Jastrzebski, Morrone,
  de~Laroussilhe, Gesmundo, Attariyan, and Gelly}]{houlsby2019parameter}
Neil Houlsby, Andrei Giurgiu, Stanislaw Jastrzebski, Bruna Morrone, Quentin
  de~Laroussilhe, Andrea Gesmundo, Mona Attariyan, and Sylvain Gelly. 2019.
\newblock Parameter-efficient transfer learning for {NLP}.
\newblock In \emph{Proceedings of the 36th International Conference on Machine
  Learning}.

\bibitem[{Howard and Ruder(2018)}]{P18-1031}
Jeremy Howard and Sebastian Ruder. 2018.
\newblock \href {http://aclweb.org/anthology/P18-1031} {Universal language
  model fine-tuning for text classification}.
\newblock In \emph{Proceedings of the 56th Annual Meeting of the Association
  for Computational Linguistics (Volume 1: Long Papers)}, pages 328--339.
  Association for Computational Linguistics.

\bibitem[{Iyyer et~al.(2015)Iyyer, Manjunatha, Boyd-Graber, and
  Daum{\'e}~III}]{iyyer2015deep}
Mohit Iyyer, Varun Manjunatha, Jordan Boyd-Graber, and Hal Daum{\'e}~III. 2015.
\newblock Deep unordered composition rivals syntactic methods for text
  classification.
\newblock In \emph{Proceedings of the 53rd Annual Meeting of the Association
  for Computational Linguistics and the 7th International Joint Conference on
  Natural Language Processing (Volume 1: Long Papers)}, volume~1, pages
  1681--1691.

\bibitem[{Jernite et~al.(2017)Jernite, Bowman, and
  Sontag}]{jernite2017discourse}
Yacine Jernite, Samuel~R. Bowman, and David Sontag. 2017.
\newblock Discourse-based objectives for fast unsupervised sentence
  representation learning.
\newblock \emph{arXiv preprint 1705.00557}.

\bibitem[{Kiela et~al.(2018)Kiela, Conneau, Jabri, and
  Nickel}]{kiela2017learning}
Douwe Kiela, Alexis Conneau, Allan Jabri, and Maximilian Nickel. 2018.
\newblock \href {http://aclweb.org/anthology/N18-1038} {Learning visually
  grounded sentence representations}.
\newblock In \emph{Proceedings of the 2018 Conference of the North American
  Chapter of the Association for Computational Linguistics: Human Language
  Technologies, Volume 1 (Long Papers)}, pages 408--418. Association for
  Computational Linguistics.

\bibitem[{Kiros et~al.(2015)Kiros, Zhu, Salakhutdinov, Zemel, Urtasun,
  Torralba, and Fidler}]{kiros2015skip}
Ryan Kiros, Yukun Zhu, Ruslan~R Salakhutdinov, Richard Zemel, Raquel Urtasun,
  Antonio Torralba, and Sanja Fidler. 2015.
\newblock Skip-{T}hought vectors.
\newblock In \emph{Advances in Neural Information Processing Systems}, pages
  3294--3302.

\bibitem[{Levesque et~al.(2011)Levesque, Davis, and
  Morgenstern}]{levesque2011winograd}
Hector~J Levesque, Ernest Davis, and Leora Morgenstern. 2011.
\newblock The {W}inograd schema challenge.
\newblock In \emph{Aaai spring symposium: Logical formalizations of commonsense
  reasoning}, volume~46, page~47.

\bibitem[{Lin et~al.(2014)Lin, Maire, Belongie, Hays, Perona, Ramanan,
  Doll{\'a}r, and Zitnick}]{lin2014microsoft}
Tsung-Yi Lin, Michael Maire, Serge Belongie, James Hays, Pietro Perona, Deva
  Ramanan, Piotr Doll{\'a}r, and C~Lawrence Zitnick. 2014.
\newblock Microsoft coco: Common objects in context.
\newblock In \emph{European conference on computer vision}, pages 740--755.
  Springer.

\bibitem[{Liu et~al.(2019)Liu, He, Chen, and Gao}]{liu2019multitask}
Xiaodong Liu, Pengcheng He, Weizhu Chen, and Jianfeng Gao. 2019.
\newblock Multi-task deep neural networks for natural language understanding.
\newblock In \emph{Proceedings of the 57th Annual Meeting of the Association
  for Computational Linguistics}. Association for Computational Linguistics.

\bibitem[{Luong et~al.(2016)Luong, Le, Sutskever, Vinyals, and
  Kaiser}]{luong2015multi}
Minh-Thang Luong, Quoc~V. Le, Ilya Sutskever, Oriol Vinyals, and Lukasz Kaiser.
  2016.
\newblock Multi-task sequence to sequence learning.
\newblock In \emph{Proceedings of the International Conference on Learning
  Representations ({ICLR})}.

\bibitem[{Marcus et~al.(1993)Marcus, Marcinkiewicz, and Santorini}]{PTB}
Mitchell~P. Marcus, Mary~Ann Marcinkiewicz, and Beatrice Santorini. 1993.
\newblock \href {http://dl.acm.org/citation.cfm?id=972470.972475} {{Building a
  Large Annotated Corpus of English: The Penn Treebank}}.
\newblock \emph{Computational Linguistics}.

\bibitem[{McCann et~al.(2017)McCann, Bradbury, Xiong, and
  Socher}]{mccann2017learned}
Bryan McCann, James Bradbury, Caiming Xiong, and Richard Socher. 2017.
\newblock Learned in translation: Contextualized word vectors.
\newblock In \emph{Advances in Neural Information Processing Systems}, pages
  6297--6308.

\bibitem[{Merity et~al.(2017)Merity, Xiong, Bradbury, and
  Socher}]{DBLP:journals/corr/MerityXBS16}
Stephen Merity, Caiming Xiong, James Bradbury, and Richard Socher. 2017.
\newblock Pointer sentinel mixture models.
\newblock In \emph{Proceedings of the International Conference on Learning
  Representations ({ICLR})}.

\bibitem[{Mou et~al.(2016)Mou, Men, Li, Xu, Zhang, Yan, and
  Jin}]{mou-EtAl:2016:P16-2}
Lili Mou, Rui Men, Ge~Li, Yan Xu, Lu~Zhang, Rui Yan, and Zhi Jin. 2016.
\newblock \href {http://anthology.aclweb.org/P16-2022} {Natural language
  inference by tree-based convolution and heuristic matching}.
\newblock In \emph{Proceedings of the 54th Annual Meeting of the Association
  for Computational Linguistics (Volume 2: Short Papers)}, pages 130--136,
  Berlin, Germany. Association for Computational Linguistics.

\bibitem[{Nie et~al.(2019)Nie, Bennett, and Goodman}]{nie2017dissent}
Allen Nie, Erin~D Bennett, and Noah~D Goodman. 2019.
\newblock Dis{S}ent: Sentence representation learning from explicit discourse
  relations.
\newblock In \emph{Proceedings of the 57th Annual Meeting of the Association
  for Computational Linguistics}. Association for Computational Linguistics.

\bibitem[{Peters et~al.(2018{\natexlab{a}})Peters, Neumann, Iyyer, Gardner,
  Clark, Lee, and Zettlemoyer}]{N18-1202}
Matthew Peters, Mark Neumann, Mohit Iyyer, Matt Gardner, Christopher Clark,
  Kenton Lee, and Luke Zettlemoyer. 2018{\natexlab{a}}.
\newblock \href {http://aclweb.org/anthology/N18-1202} {Deep contextualized
  word representations}.
\newblock In \emph{Proceedings of the 2018 Conference of the North American
  Chapter of the Association for Computational Linguistics: Human Language
  Technologies, Volume 1 (Long Papers)}, pages 2227--2237. Association for
  Computational Linguistics.

\bibitem[{Peters et~al.(2019)Peters, Ruder, and Smith}]{peters2019tune}
Matthew Peters, Sebastian Ruder, and Noah~A. Smith. 2019.
\newblock To tune or not to tune? adapting pretrained representations to
  diverse tasks.
\newblock \emph{arXiv preprint 1903.05987}.

\bibitem[{Peters et~al.(2018{\natexlab{b}})Peters, Neumann, Zettlemoyer, and
  Yih}]{peters2018dissecting}
Matthew~E. Peters, Mark Neumann, Luke Zettlemoyer, and Wen-tau Yih.
  2018{\natexlab{b}}.
\newblock Dissecting contextual word embeddings: Architecture and
  representation.
\newblock In \emph{Proceedings of the 2018 Conference on Empirical Methods in
  Natural Language Processing ({EMNLP})}.

\bibitem[{Phang et~al.(2018)Phang, Févry, and Bowman}]{phang2018sentence}
Jason Phang, Thibault Févry, and Samuel~R. Bowman. 2018.
\newblock Sentence encoders on {STILTs}: Supplementary training on intermediate
  labeled-data tasks.
\newblock \emph{arXiv preprint 1811.01088}.

\bibitem[{Poliak et~al.(2018)Poliak, Haldar, Rudinger, Hu, Pavlick, White, and
  Van~Durme}]{poliak2018towards}
Adam Poliak, Aparajita Haldar, Rachel Rudinger, J~Edward Hu, Ellie Pavlick,
  Aaron~Steven White, and Benjamin Van~Durme. 2018.
\newblock Towards a unified natural language inference framework to evaluate
  sentence representations.
\newblock \emph{arXiv preprint 1804.08207}.

\bibitem[{Radford et~al.(2018)Radford, Narasimhan, Salimans, and
  Sutskever}]{radford2018improving}
Alec Radford, Karthik Narasimhan, Tim Salimans, and Ilya Sutskever. 2018.
\newblock \href {https://blog.openai.com/language-unsupervised/} {Improving
  language understanding by generative pre-training}.
\newblock Unpublished manuscript accessible via the OpenAI Blog.

\bibitem[{Radford et~al.(2019)Radford, Wu, Child, Luan, Amodei, and
  Sutskever}]{radford2019language}
Alec Radford, Jeffrey Wu, Rewon Child, David Luan, Dario Amodei, and Ilya
  Sutskever. 2019.
\newblock \href
  {https://d4mucfpksywv.cloudfront.net/better-language-models/language-models.pdf}
  {Improving language understanding by generative pre-training}.
\newblock Unpublished manuscript accessible via the OpenAI Blog.

\bibitem[{Rajpurkar et~al.(2018)Rajpurkar, Jia, and Liang}]{rajpurkar2018know}
Pranav Rajpurkar, Robin Jia, and Percy Liang. 2018.
\newblock \href {http://www.aclweb.org/anthology/P18-2124} {Know what you don't
  know: Unanswerable questions for squad}.
\newblock In \emph{Proceedings of the 56th Annual Meeting of the Association
  for Computational Linguistics (Volume 2: Short Papers)}, pages 784--789,
  Melbourne, Australia. Association for Computational Linguistics.

\bibitem[{Rajpurkar et~al.(2016)Rajpurkar, Zhang, Lopyrev, and
  Liang}]{rajpurkar2016squad}
Pranav Rajpurkar, Jian Zhang, Konstantin Lopyrev, and Percy Liang. 2016.
\newblock \href {https://doi.org/10.18653/v1/D16-1264} {{SQ}u{AD}: 100,000+
  questions for machine comprehension of text}.
\newblock In \emph{Proceedings of the 2016 Conference on Empirical Methods in
  Natural Language Processing}, pages 2383--2392. Association for Computational
  Linguistics.

\bibitem[{Reddi et~al.(2018)Reddi, Kale, and Kumar}]{j.2018on}
Sashank~J. Reddi, Satyen Kale, and Sanjiv Kumar. 2018.
\newblock On the convergence of {Adam} and beyond.
\newblock In \emph{Proceedings of the International Conference on Learning
  Representations ({ICLR})}.

\bibitem[{Seo et~al.(2017)Seo, Kembhavi, Farhadi, and
  Hajishirzi}]{seo2016bidirectional}
Minjoon Seo, Aniruddha Kembhavi, Ali Farhadi, and Hannaneh Hajishirzi. 2017.
\newblock Bidirectional attention flow for machine comprehension.
\newblock In \emph{Proceedings of the International Conference on Learning
  Representations ({ICLR})}.

\bibitem[{Socher et~al.(2013)Socher, Perelygin, Wu, Chuang, Manning, Ng, and
  Potts}]{socher2013recursive}
Richard Socher, Alex Perelygin, Jean Wu, Jason Chuang, Christopher~D Manning,
  Andrew Ng, and Christopher Potts. 2013.
\newblock Recursive deep models for semantic compositionality over a sentiment
  treebank.
\newblock In \emph{Proceedings of the 2013 conference on empirical methods in
  natural language processing}, pages 1631--1642.

\bibitem[{Stickland and Murray(2019)}]{stickl2019bert}
Asa~Cooper Stickland and Iain Murray. 2019.
\newblock {BERT} and {PALs}: Projected attention layers for efficient
  adaptation in multi-task learning.
\newblock In \emph{Proceedings of the 36th International Conference on Machine
  Learning}.

\bibitem[{Subramanian et~al.(2018)Subramanian, Trischler, Bengio, and
  Pal}]{subramanian2018large}
Sandeep Subramanian, Adam Trischler, Yoshua Bengio, and Christopher~J. Pal.
  2018.
\newblock Learning general purpose distributed sentence representations via
  large scale multi-task learning.
\newblock In \emph{Proceedings of the International Conference on Learning
  Representations ({ICLR})}.

\bibitem[{Tang et~al.(2017)Tang, Jin, Fang, Wang, and de~Sa}]{W17-2625}
Shuai Tang, Hailin Jin, Chen Fang, Zhaowen Wang, and Virginia de~Sa. 2017.
\newblock \href {http://aclweb.org/anthology/W17-2625} {Rethinking
  {S}kip-thought: A neighborhood based approach}.
\newblock In \emph{Proceedings of the 2nd Workshop on Representation Learning
  for {NLP}}, pages 211--218. Association for Computational Linguistics.

\bibitem[{Tenney et~al.(2019{\natexlab{a}})Tenney, Das, and
  Pavlick}]{tenney2019bert}
Ian Tenney, Dipanjan Das, and Ellie Pavlick. 2019{\natexlab{a}}.
\newblock {BERT} rediscovers the classical nlp pipeline.
\newblock In \emph{Proceedings of the 57th Annual Meeting of the Association
  for Computational Linguistics}. Association for Computational Linguistics.

\bibitem[{Tenney et~al.(2019{\natexlab{b}})Tenney, Xia, Chen, Wang, Poliak,
  McCoy, Kim, Durme, Bowman, Das, and Pavlick}]{tenney2018what}
Ian Tenney, Patrick Xia, Berlin Chen, Alex Wang, Adam Poliak, R~Thomas McCoy,
  Najoung Kim, Benjamin~Van Durme, Sam Bowman, Dipanjan Das, and Ellie Pavlick.
  2019{\natexlab{b}}.
\newblock \href {https://openreview.net/forum?id=SJzSgnRcKX} {What do you learn
  from context? probing for sentence structure in contextualized word
  representations}.
\newblock In \emph{Proceedings of the International Conference on Learning
  Representations ({ICLR})}.

\bibitem[{Vaswani et~al.(2017)Vaswani, Shazeer, Parmar, Uszkoreit, Jones,
  Gomez, Kaiser, and Polosukhin}]{vaswani2017attention}
Ashish Vaswani, Noam Shazeer, Niki Parmar, Jakob Uszkoreit, Llion Jones,
  Aidan~N Gomez, {\L}ukasz Kaiser, and Illia Polosukhin. 2017.
\newblock Attention is all you need.
\newblock In \emph{Advances in Neural Information Processing Systems}, pages
  6000--6010.

\bibitem[{Wang et~al.(2019)Wang, Singh, Michael, Hill, Levy, and
  Bowman}]{wang2018glue}
Alex Wang, Amanpreet Singh, Julian Michael, Felix Hill, Omer Levy, and
  Samuel~R. Bowman. 2019.
\newblock {GLUE}: A multi-task benchmark and analysis platform for natural
  language understanding.
\newblock In \emph{Proceedings of the International Conference on Learning
  Representations ({ICLR})}.

\bibitem[{Warstadt et~al.(2018)Warstadt, Singh, and
  Bowman}]{warstadt2018neural}
Alex Warstadt, Amanpreet Singh, and Samuel~R. Bowman. 2018.
\newblock Neural network acceptability judgments.
\newblock \emph{arXiv preprint 1805.12471}.

\bibitem[{White et~al.(2017)White, Rastogi, Duh, and
  Van~Durme}]{white2017inference}
Aaron~Steven White, Pushpendre Rastogi, Kevin Duh, and Benjamin Van~Durme.
  2017.
\newblock Inference is everything: Recasting semantic resources into a unified
  evaluation framework.
\newblock In \emph{Proceedings of the Eighth International Joint Conference on
  Natural Language Processing (Volume 1: Long Papers)}, volume~1, pages
  996--1005.

\bibitem[{Wieting and Kiela(2019)}]{wieting2019training}
John Wieting and Douwe Kiela. 2019.
\newblock No training required: Exploring random encoders for sentence
  classification.
\newblock In \emph{Proceedings of the International Conference on Learning
  Representations ({ICLR})}.

\bibitem[{Williams et~al.(2018)Williams, Nangia, and
  Bowman}]{DBLP:journals/corr/WilliamsNB17}
Adina Williams, Nikita Nangia, and Samuel Bowman. 2018.
\newblock \href {http://aclweb.org/anthology/N18-1101} {A broad-coverage
  challenge corpus for sentence understanding through inference}.
\newblock In \emph{Proceedings of the 2018 Conference of the North American
  Chapter of the Association for Computational Linguistics: Human Language
  Technologies, Volume 1 (Long Papers)}, pages 1112--1122. Association for
  Computational Linguistics.

\bibitem[{Wu et~al.(2016)Wu, Schuster, Chen, Le, Norouzi, Macherey, Krikun,
  Cao, Gao, Macherey et~al.}]{wu2016google}
Yonghui Wu, Mike Schuster, Zhifeng Chen, Quoc~V Le, Mohammad Norouzi, Wolfgang
  Macherey, Maxim Krikun, Yuan Cao, Qin Gao, Klaus Macherey, et~al. 2016.
\newblock {G}oogle's neural machine translation system: {B}ridging the gap
  between human and machine translation.
\newblock \emph{arXiv preprint 1609.08144}.

\bibitem[{Yang et~al.(2018)Yang, Yuan, Cer, Kong, Constant, Pilar, Ge, Sung,
  Strope, and Kurzweil}]{yang2018learning}
Yinfei Yang, Steve Yuan, Daniel Cer, Sheng-Yi Kong, Noah Constant, Petr Pilar,
  Heming Ge, Yun-hsuan Sung, Brian Strope, and Ray Kurzweil. 2018.
\newblock \href {http://www.aclweb.org/anthology/W18-3022} {Learning semantic
  textual similarity from conversations}.
\newblock In \emph{Proceedings of The Third Workshop on Representation Learning
  for NLP}, pages 164--174, Melbourne, Australia. Association for Computational
  Linguistics.

\bibitem[{Yogatama et~al.(2019)Yogatama, de~Masson~d'Autume, Connor, Kocisky,
  Chrzanowski, Kong, Lazaridou, Ling, Yu, Dyer, and
  Blunsom}]{yogatama2019learning}
Dani Yogatama, Cyprien de~Masson~d'Autume, Jerome Connor, Tomas Kocisky, Mike
  Chrzanowski, Lingpeng Kong, Angeliki Lazaridou, Wang Ling, Lei Yu, Chris
  Dyer, and Phil Blunsom. 2019.
\newblock Learning and evaluating general linguistic intelligence.
\newblock \emph{arXiv preprint 1901.11373}.

\bibitem[{Zhang and Bowman(2018)}]{zhang2018lessons}
Kelly Zhang and Samuel~R. Bowman. 2018.
\newblock Language modeling teaches you more syntax than translation does:
  Lessons learned through auxiliary task analysis.
\newblock \emph{arXiv preprint 1809.10040}.

\bibitem[{Zhang et~al.(2017)Zhang, Rudinger, Duh, and
  Van~Durme}]{zhang2017ordinal}
Sheng Zhang, Rachel Rudinger, Kevin Duh, and Benjamin Van~Durme. 2017.
\newblock Ordinal common-sense inference.
\newblock \emph{Transactions of the Association of Computational Linguistics},
  5(1):379--395.

\end{thebibliography}
